\documentclass[runningheads]{llncs}
\usepackage[T1]{fontenc}
\usepackage[most]{tcolorbox}
\usepackage[inline]{enumitem}
\makeatletter
\renewcommand\paragraph{\@startsection{paragraph}{4}{\z@}%
  {0.6ex \@plus 0.2ex \@minus 0.2ex}
  {-0.8em}
  {\normalfont\normalsize\bfseries}}
\makeatother

\usepackage[numbers,sort&compress]{natbib}
\usepackage{graphicx} 
\usepackage{caption}  
\usepackage{float}    
\usepackage{booktabs}
\usepackage{tabularx}
\usepackage{makecell}
\usepackage{multirow}
\usepackage{amsmath}
\usepackage{wrapfig}
\usepackage{enumitem}
\usepackage{threeparttable}
\usepackage{siunitx}
\usepackage{adjustbox}
\usepackage{tablefootnote}
\usepackage[a4paper, left=1.5cm, right=1.5cm, top=2cm, bottom=2cm]{geometry}
\usepackage{url}

\usepackage{xcolor}
\definecolor{Agreen}{RGB}{0,100,0}      
\definecolor{Cred}{RGB}{139,0,0}        

\usepackage{pifont}
\newcommand{\cmark}{\ding{51}} 
\newcommand{\xmark}{\ding{55}} 

\usepackage{listings}
\usepackage{xcolor}

\lstdefinestyle{promptstyle}{
  basicstyle=\ttfamily\footnotesize, 
  breaklines=true,
  breakatwhitespace=false,
  columns=fullflexible,
  keepspaces=true,
  showstringspaces=false,
  frame=single,
  framerule=0.4pt,
  xleftmargin=0.5em,
  xrightmargin=0.5em
}

\begin{document}
\title{Event-Centric Human Value Understanding in News-Domain Texts: An Actor-Conditioned, Multi-Granularity Benchmark}
%
%
\author{
    Yao Wang\inst{1} \and
    Xin Liu\inst{2} \and
    Zhuochen Liu\inst{1} \and
    Jiankang Chen\inst{1} \and
    Adam Jatowt\inst{3} \and
    Kyoungsook Kim\inst{2} \and
    Noriko Kando\inst{4} \and
    Haitao Yu\inst{5}\thanks{Corresponding author.}
}

\authorrunning{Y. Wang et al.}

\institute{
Graduate School of Comprehensive Human Sciences, University of Tsukuba, Tsukuba, Japan
\email{oh.gyou.tkb\_gf@u.tsukuba.ac.jp, s2426082@u.tsukuba.ac.jp, s2526100@u.tsukuba.ac.jp}
\and
 Intelligent Platforms Research Institute, National Institute of Advanced Industrial Science and Technology, Tokyo, Japan
\email{xin.liu@aist.go.jp, ks.kim@aist.go.jp}
\and
Digital Science Center, University of Innsbruck, Innsbruck, Austria
\email{adam.jatowt@uibk.ac.at}
\and
National Institute of Informatics, Tokyo, Japan
\email{kando@nii.ac.jp}
\and
Institute of Library, Information and Media Science, University of Tsukuba, Tsukuba, Japan
\email{yuhaitao@slis.tsukuba.ac.jp}
}

\maketitle              
\begin{abstract}
Existing human value datasets do not directly support value understanding in factual news: many are actor-agnostic, rely on isolated utterances or synthetic scenarios, and lack explicit event structure or value direction.
We present \textbf{NEVU} (\textbf{N}ews \textbf{E}vent-centric \textbf{V}alue \textbf{U}nderstanding), a benchmark for \emph{actor-conditioned}, \emph{event-centric}, and \emph{direction-aware} human value recognition in factual news.
NEVU evaluates whether models can identify value cues, attribute them to the correct actor, and determine value direction from grounded evidence.
Built from 2{,}865 English news articles, NEVU organizes annotations at four semantic unit levels (\textbf{Subevent}, \textbf{behavior-based composite event}, \textbf{story-based composite event}, and \textbf{Article}) and labels \mbox{(unit, actor)} pairs for fine-grained evaluation across local and composite contexts.
The annotations are produced through an LLM-assisted pipeline with staged verification and targeted human auditing.
Using a hierarchical value space with \textbf{54} fine-grained values and \textbf{20} coarse-grained categories, NEVU covers 45{,}793 unit--actor pairs and 168{,}061 directed value instances.
We provide unified baselines for proprietary and open-source LLMs, and find that lightweight adaptation (LoRA) consistently improves open-source models, showing that although NEVU is designed primarily as a benchmark, it also supports supervised adaptation beyond prompting-only evaluation. Data availability is described in Appendix~\ref{app:data_code_availability}.

\end{abstract}

\section{Introduction}
\label{sec:intro}

\begin{wrapfigure}{r}[0pt]{0.60\textwidth}
  \centering
  \vspace{-30pt}
  \includegraphics[width=\linewidth, trim=3mm 85mm 120mm 20mm, clip]{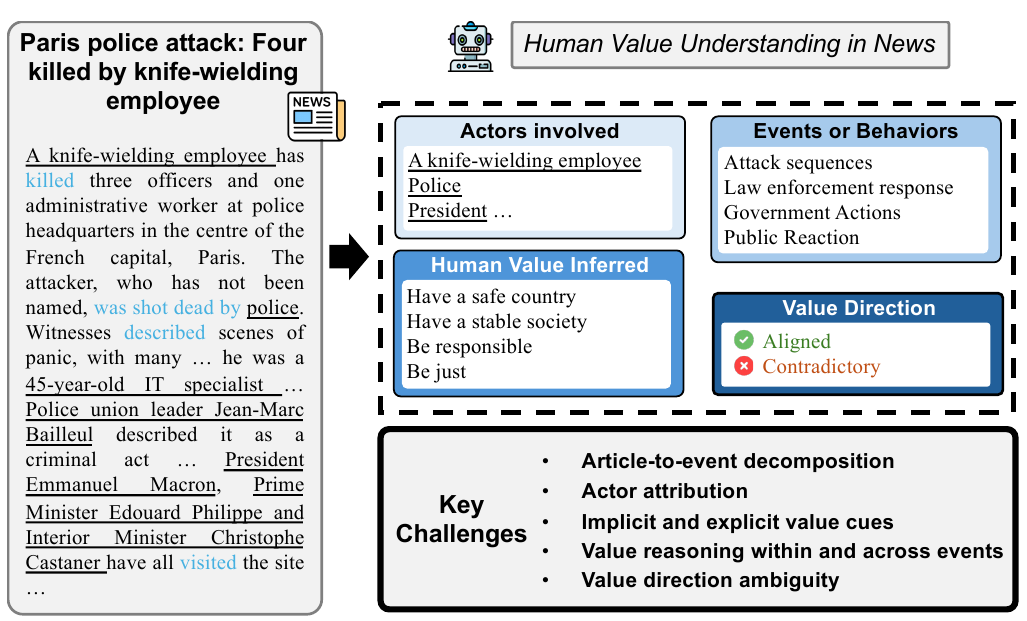}
  \caption{\small Human Value Understanding in News: Core Components and Challenges.}
  \label{fig:nevu_intro}
  \vspace{-15pt}
\end{wrapfigure}

Human values are commonly understood as normative principles that guide how individuals and societies judge actions and outcomes \cite{schwartz2012refining, rokeach1973nature, brown2002life, haerpfer2020world}. 
As large language models (LLMs) are increasingly used to read, summarize, and explain news \cite{hu2024bad, pratelli2025evaluation, zhang2025systematic, yang2025accuracy, cao2025survey}, they do not merely generate fluent text, but also shape how information is interpreted. 
In this setting, factual accuracy alone is not sufficient. 
A system may correctly report what happened while still misrepresenting the value-laden aspects of the news, for example by assigning blame or credit to the wrong actors or framing events in ways that reflect model priors rather than source evidence. 
This creates a need to evaluate whether models can handle value-relevant aspects of news in an evidence-grounded and actor-sensitive way.

News is a particularly challenging domain for human value understanding. 
Unlike short arguments, dialogues, elicited value statements, or synthetic moral scenarios, factual news reports describe real-world social processes involving multiple actors, actions, institutional constraints, and downstream consequences. 
Rather than recognizing explicitly stated opinions or normative judgments, models must infer values from evidence-grounded descriptions of behaviors, interactions, and outcomes. 
As illustrated in Figure~\ref{fig:nevu_intro}, this requires identifying \emph{who} is involved, \emph{what} events and behaviors are described, \emph{which} human values are implicated, and \emph{whether} those values are expressed in an aligned or contradictory direction. 
These judgments are difficult because value cues may be explicit or implicit, distributed within or across events, and intertwined with actor attribution and narrative framing \cite{entman1993framing, devreese2005news, semetko2000framing}. 
As a result, article-level labeling alone is often too coarse to reveal whether a model has missed relevant evidence, confused actors, or reversed value direction. 
What is needed is a benchmark that evaluates value understanding in factual news through explicit event and narrative structure, rather than as a single document-level judgment.

\begin{wrapfigure}{r}[0pt]{0.60\textwidth}
  \centering
  \vspace{-20pt}
  \includegraphics[width=\linewidth, trim=3mm 77mm 80mm 20mm, clip]{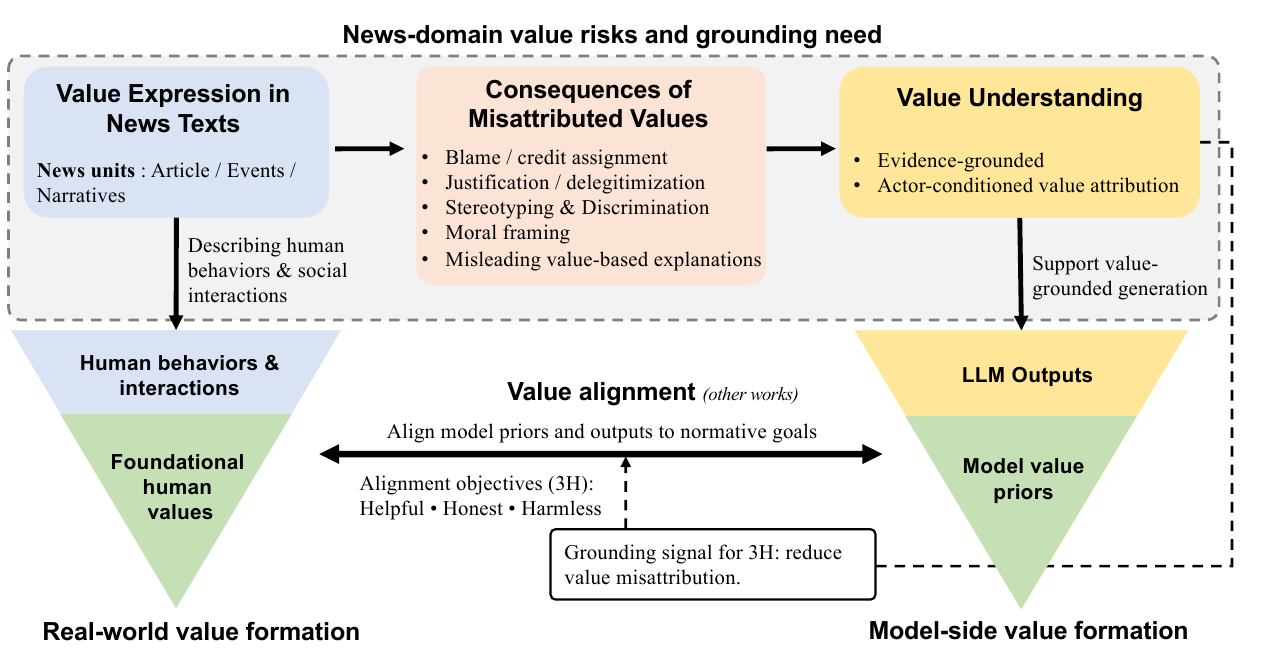}
  \caption{\small Evidence-Grounded Value Understanding for News-Domain Texts.}
  \label{fig:intro_positioning}
  \vspace{-10pt}
\end{wrapfigure}

These challenges are only partially addressed by existing value-related research in NLP.
One line of work studies human value recognition in arguments, dialogues, and other non-news text types \cite{kiesel2022identifying, qiu2022valuenet, kiesel2023semeval, sorensen2024value}.
Although valuable, these resources typically focus on isolated utterances or short texts, are often actor-agnostic in explicit social attribution, and do not model value judgments over structured real-world events in factual news.
Another line of work studies alignment and controllability, asking whether model outputs can be guided toward desired normative preferences or value expressions \cite{askell2021general, bai2022constitutional, ji2023beavertails, guo2025counterfactual, yang2025constructing, rezapour2025tales, su2025understanding, yao2024clave, trager2025mftcxplain, chakraborty2025structured}.
Figure~\ref{fig:intro_positioning} provides a schematic view of this landscape.
In contrast to output-side alignment, we focus on the input-side problem of identifying which values are expressed in source news, by which actors, and in what direction. However, no existing benchmark for factual news jointly covers \emph{event-centric decomposition}, \emph{actor-conditioned attribution}, \emph{multi-granularity analysis}, and \emph{direction-aware} value recognition.

To fill this gap, we introduce \textbf{NEVU} (\textbf{N}ews \textbf{E}vent-centric \textbf{V}alue \textbf{U}nderstanding), a benchmark for fine-grained human value understanding in factual news. 
NEVU is designed primarily for evaluation, while its structured annotations also support the study of learnability beyond prompting-only settings. Within NEVU, each article is organized into four semantic levels (\textit{Subevent}, \textit{behavior-based composite event (BCE)}, \textit{story-based composite event (SCE)}, and \textit{Article}), with directed human value annotations defined over \mbox{(unit, actor)} pairs.
This hierarchy supports value understanding across multiple semantic scopes: \textit{Subevents} localize value-relevant evidence, while \textit{BCEs} and \textit{SCEs} organize related subevents into relatively more complete behavior chains and story episodes. 
These composite units provide broader contextual grounding beyond isolated subevents, which is useful when value implications become clearer only from linked actions, responses, and consequences viewed together. 
Based on this structure, we formulate \textbf{Human Value Recognition} in news as an actor-conditioned, direction-aware, and multi-granularity benchmark task. 
We adopt the hierarchical value taxonomy of Kiesel et al.~\cite{kiesel2022identifying}, using both Level-1 (54 fine-grained values) and Level-2 (20 coarse-grained categories); detailed definitions are provided in Appendix~\ref{app:value_hierarchy}. 
Problem formulation and dataset construction details are given in Sections~\ref{sec:problem} and~\ref{sec:nevu_pipeline}.

We summarize our main contributions as follows:

\begin{itemize}[label=\textbullet]

    \item We formalize \textbf{Human Value Recognition} in factual news as an evidence-grounded benchmark task over \textit{Article}, \textit{Subevent}, \textit{BCE}, and \textit{SCE} units, enabling actor-conditioned and direction-aware evaluation across multiple semantic granularities.
    
    \item We present \textbf{NEVU}, the first benchmark for \emph{event-centric}, \emph{actor-conditioned}, and \emph{direction-aware} human value understanding in factual news, comprising \textbf{2,865} articles, \textbf{25,753} subevents, \textbf{9,216} BCEs, \textbf{7,959} SCEs, \textbf{45,793} unit--actor pairs, and \textbf{168,061} directed value instances.

    \item We establish a systematic benchmark protocol and evaluate proprietary and open-source LLMs under a shared hierarchical value space (Level-1 \textbf{54} / Level-2 \textbf{20}), enabling direct comparison of model performance on fine-grained and direction-aware news-domain value recognition.

    \item We further show that NEVU supports learnability beyond prompting-only evaluation: lightweight supervised adaptation (LoRA) consistently improves open-source LLMs on fine-grained value and direction recognition, providing complementary evidence of the benchmark's utility.
\end{itemize}

\begin{table}[t]
\centering
\scriptsize
\setlength{\tabcolsep}{3.0pt}
\renewcommand{\arraystretch}{1.10}

\caption{Comparison of human value datasets by data source, text type, and annotation properties.}
\label{tab:value_datasets}

\begin{adjustbox}{max width=\columnwidth}
\begin{tabular}{l c l l c r c c c}
\toprule
\textbf{Dataset} & \textbf{Year} & \textbf{Data source} & \textbf{Text type} &
\textbf{Actor} & \textbf{Size} &
\textbf{Card.} & \textbf{Relation} & \textbf{Open} \\
\midrule
Kiesel22-Args \cite{kiesel2022identifying} & 2022 & IBM-ArgQ-Rank-30kArgs \cite{gretz2019largescaledatasetargumentquality} & Argument & \xmark & 5.3k
& Multi & Undir. & Yes \\

ValueNet \cite{qiu2022valuenet} & 2022 & EmpatheticDialogues \cite{rashkin-etal-2019-towards} & Dialogue & \xmark & 21.4k
& Single & Dir. & Yes \\

ValueEval \cite{kiesel2023semeval} & 2023 & Touch\'e23-ValueEval \cite{mirzakhmedova2024touche23} & Argument & \xmark & 9.3k
& Multi & Undir. & Yes \\

ValueFULCRA \cite{yao2024value} & 2023 & Red Teaming \cite{perez-etal-2022-red,bai2022traininghelpfulharmlessassistant} & LLM answers & \xmark & 20k
& Multi & Dir. & No \\

ValueBench \cite{ren2024valuebench} & 2024 & Psychometric inventories \cite{fraser-etal-2022-moral,karra2022estimating,caron2022identifying,li2023does,miotto2022gpt,rao-etal-2023-chatgpt,jiang2023evaluating,song2023have,zhang2023measuring,zhang2023heterogeneous,pan2023llms,serapio2023personality,ganesan2023systematic,huang-etal-2024-reliability,abdulhai2024moral,simmons2023moral,scherrer2023evaluating,bodroza2023personality,la2025open} & Questionnaire item & \xmark & 2k
& Single & Dir. & Yes \\

ValuePrism \cite{sorensen2024value} & 2024 & Delphi scenarios \cite{jiang2021can} & Situation vignette & \xmark & 31k
& Multi & Dir. & Yes \\

VVALUES \cite{10.1016/j.neucom.2025.130170} & 2025 &
TikTok/Twitter\tablefootnote{\url{https://www.tiktok.com/}, \url{https://x.com/}}
& Video (+text) & \xmark & 5.1k
& Multi & Dir. & No \\

\midrule
\textbf{NEVU (ours)} & 2026 & Uknow \cite{gong2024uknow} + social news (Table~\ref{tab:app_source_domains}) & \textbf{Multi-level event units} &
\cmark & \textbf{46.6k}
& Multi & Dir. & Yes \\
\bottomrule
\end{tabular}
\end{adjustbox}

\vspace{1pt}
\footnotesize
\raggedright
\noindent \textit{Note.} Size is the dataset scale reported in the original paper; Card. indicates whether each instance is annotated with a single value label or multiple value labels; Relation indicates whether annotations distinguish value direction (Dir.: aligned vs.\ contradictory) or only mark value presence (Undir.); Open indicates whether the dataset is publicly released. For NEVU, Size counts total semantic units across its four levels (46.6k in total).
\end{table}

\section{Related Work}
\label{sec:related}

\subsection{Human Value Benchmarks Beyond Factual News}

Human value research has a long tradition in psychology, sociology, and political communication, where values are commonly understood as enduring principles that shape how individuals and societies evaluate actions, choices, and social life \cite{schwartz2012refining, rokeach1973nature, brown2002life, haerpfer2020world}. 
In NLP, these foundations have been operationalized through benchmark datasets that make value recognition measurable over text. 
Early work by \cite{kiesel2022identifying} introduced a hierarchical value taxonomy and annotated argumentation data, establishing human value recognition as a supervised NLP task. 
Subsequent work extended value-related resources to dialogues and interactive text \cite{qiu2022valuenet, shen2025valuecompass}, psychometric-style questionnaire items \cite{ren2024valuebench}, synthetic moral or value scenarios \cite{lourie2021scruplescorpuscommunityethical, marcuzzo2025morables, chen2025mova}, and broader value-oriented evaluation settings \cite{sorensen2024value}.

As summarized in Table~\ref{tab:value_datasets}, existing value benchmarks differ substantially in source domain, annotation unit, and label structure. 
Some treat value prediction as undirected relevance or presence tagging, without distinguishing whether a value is expressed in an aligned or contradictory direction \cite{kiesel2022identifying, kiesel2023semeval}, while others include directional distinctions over dialogues, scenarios, or multimodal inputs \cite{qiu2022valuenet, sorensen2024value, 10.1016/j.neucom.2025.130170}. 
These datasets provide important foundations for value-related NLP research, but they typically focus on isolated utterances, short texts, questionnaire items, or synthetic situations rather than factual news reports. 
Moreover, most do not explicitly support \emph{actor-conditioned} value attribution over structured real-world events in factual news.
Against this background, NEVU shifts the focus to factual news and jointly supports actor-conditioned attribution, direction-aware value labels, and multi-granularity event structure.

\subsection{Event-Centric Structure as a Missing Dimension in Value Benchmarking}

Event extraction, structured event representations, and related work on event relations have long supported news information extraction, summarization, and reasoning \cite{huang2024textee, chen2020event, liu2025utilizing, guan2022event, jin2022event, zheng2024comprehensive}. 
A central motivation in this line of work is that news articles often interleave multiple actors, actions, and developments, making article-level analysis alone too coarse for localizable reasoning \cite{li2021future, chen2020event, liu2025utilizing, keith2023survey}. 
Such structure is particularly useful in news, where the same article may contain parallel developments, temporally distributed actions, and multiple actor roles \cite{zhang2024analyzing, ayyubi2024beyond}.

This motivation is especially important for human value understanding, where value evidence often emerges across event progressions and actor interactions (Figure~\ref{fig:nevu_intro}).
Value-relevant evidence in news is often distributed across behaviors, interactions, responsibility assignments, institutional responses, and downstream consequences, rather than concentrated in a single sentence or local span. 
As a result, fine-grained subevent decomposition is useful but not always sufficient: broader contextual units may be needed when value-relevant evidence is distributed across linked actions, responses, and consequences \cite{chen2023cheer, zheng2024comprehensive, huang2024textee}. 
However, although prior event-centric work provides structured representations for news, it does not directly benchmark human values, actor-conditioned attribution, or directional value judgments. 
NEVU addresses this missing dimension by annotating directed human values over multi-granularity event units, including localized subevents and broader composite units that provide richer contextual grounding for value attribution.

\subsection{Actor Attribution and Value Expression in News}

Communication research has long shown that news shapes audience understanding through framing, attribution, and narrative organization \cite{entman1993framing, devreese2005news, semetko2000framing}. 
In factual news, value cues are often indirect: they may be implied through descriptions of behaviors, institutional constraints, policy choices, responsibility assignments, and consequences rather than stated as explicit opinions or moral claims. 
News reports also frequently combine quoted speech, reported claims, and differing accounts from multiple actors, which makes interpretation sensitive to who is speaking, who is acting, and whose perspective is foregrounded.

These properties make actor attribution central to value understanding in news. 
Models must often determine which actor expresses or instantiates a value-relevant position, whether a described action aligns with or contradicts a value, and which evidence in the report supports that judgment \cite{zhang2022directquote, glockner2025neoqa}. 
Related NLP work has studied stance detection, bias analysis, reframing, and interpretive differences in news \cite{zhang2024survey, nguyen2024news, knutson2024news, kmainasi2025llamalens}. 
These directions highlight the importance of attribution and perspective, but they do not directly provide benchmark resources for \emph{human value understanding} in news. 
In particular, benchmarks that operationalize value understanding as \emph{actor-conditioned}, \emph{direction-aware} attribution over factual news remain limited. 
NEVU addresses this gap by binding directed value labels to \mbox{(unit, actor)} pairs under explicit event structure.

\subsection{Output-Side Alignment vs.\ Input-Side Value Understanding}

Value-related research on large language models has increasingly focused on normative alignment, controllability, and safety-oriented generation \cite{moore-etal-2024-large, norhashim2024measuring, ozeki-etal-2025-normative, sachdeva2025normative, yao2023instructions, wang2023aligning}. 
A closely related line of work studies whether model outputs can be guided toward desired normative preferences or value expressions \cite{askell2021general, bai2022constitutional, ji2023beavertails, guo2025counterfactual, yang2025constructing, rezapour2025tales, su2025understanding, yao2024clave, trager2025mftcxplain, chakraborty2025structured}. 
This literature primarily addresses the \emph{output-side} problem of steering model behavior toward preferred values, safety constraints, or interaction principles such as helpfulness, honesty, and harmlessness \cite{askell2021general}.

By contrast, NEVU targets the \emph{input-side} problem of human value understanding: identifying which values are expressed in source news, by which actors, and in what direction, under grounded evidence and explicit event structure. 
Input-side human value recognition has been benchmarked on arguments, dialogues, and scenario-based texts \cite{kiesel2022identifying, qiu2022valuenet, kiesel2023semeval, sorensen2024value}, but factual news introduces additional challenges, including multiple actors, event-structured context, and value cues distributed across related developments \cite{zhang2024moka, becker2025moralization}. 
NEVU therefore complements output-side alignment research by providing a benchmark for evidence-grounded, actor-conditioned, and direction-aware value recognition in factual news, as summarized in Figure~\ref{fig:intro_positioning}.

\section{Problem Formulation}
\label{sec:problem}

We formulate human value understanding in news as an evidence-grounded prediction problem over event-structured semantic units.
Unlike document-level value tagging, the goal is to identify \emph{which} values are expressed, \emph{by which actor}, \emph{at which semantic granularity}, and \emph{in what direction} (\emph{aligned} vs.\ \emph{contradictory}), while keeping predictions reviewable against localized evidence.
Let $g \in \mathcal{G}$ denote a news article.
For each article, we define a hierarchy of semantic units

\begin{equation}
\label{eq:unit-set}
\mathcal{U}^{(g)}=\{u^{(g)}_{\mathrm{art}}\}\cup \mathcal{E}^{(g)}\cup \mathcal{B}^{(g)}\cup \mathcal{N}^{(g)}.
\end{equation}

where $u^{(g)}_{\mathrm{art}}$ is the article-level unit, $\mathcal{E}^{(g)}$ is the set of subevents, $\mathcal{B}^{(g)}$ is the set of BCEs, and $\mathcal{N}^{(g)}$ is the set of SCEs.
We also define a set of major social actors $\mathcal{A}^{(g)}$ for each article.

Given a semantic unit $u \in \mathcal{U}^{(g)}$ and a target actor $a \in \mathcal{A}^{(g)}$, the task is to predict which human values are expressed with respect to the actor's described behaviors, interactions, constraints, or outcomes in $u$.
Let $V$ denote the human value label space.
For each instance $(g,u,a)$, the gold annotation is represented as two subsets
\begin{equation}
\label{eq:dir-label-sets}
Y^{+}_{g,u,a}\subseteq V, \qquad Y^{-}_{g,u,a}\subseteq V.
\end{equation}
where $Y^{+}_{g,u,a}$ contains values expressed in an aligned direction and $Y^{-}_{g,u,a}$ contains values expressed in a contradictory direction.
The prediction target is thus a directed multi-label output over \mbox{(unit, actor)} pairs.
In \textbf{ECHV}, we instantiate this formulation as the benchmark task of \textbf{Human Value Recognition}; the dataset provides event-structured units, normalized actors, and directed value annotations to enable fine-grained attribution analysis and direction-aware evaluation.
We next describe how the dataset is constructed.

\section{Dataset Construction}
\label{sec:nevu_pipeline}

\begin{wrapfigure}{r}{0.48\textwidth}
\vspace{-16pt}
\centering
\includegraphics[width=0.46\textwidth, trim=3mm 85mm 150mm 22mm, clip]{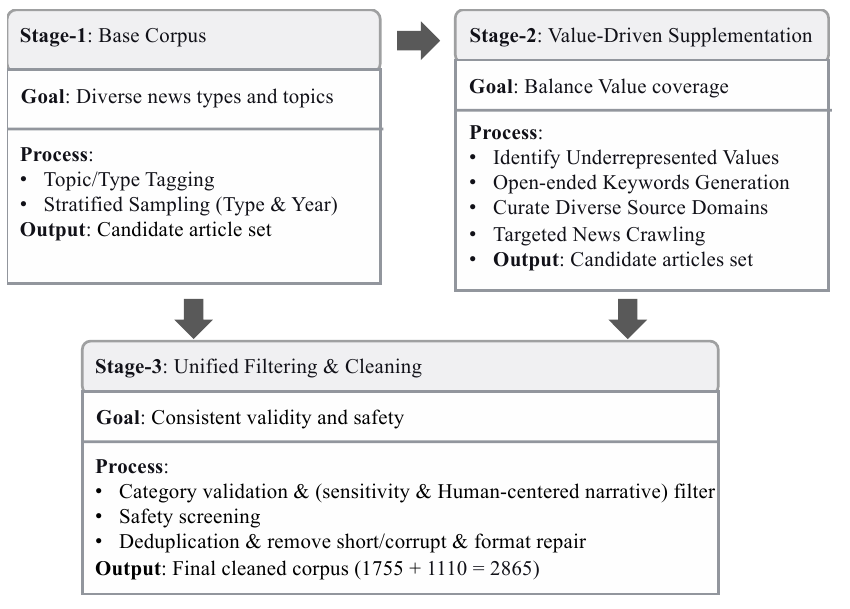}
\vspace{-6pt}
\caption{\footnotesize Phase-1 pipeline for corpus construction.}
\label{fig:phase1_flow}
\vspace{-60pt}
\end{wrapfigure}

To instantiate the problem formulation in Section~\ref{sec:problem}, we construct NEVU through a three-phase pipeline covering corpus collection, event-centric knowledge base construction, and actor-conditioned human value annotation.
\subsection{Phase-1: Data Collection}
\label{sec:phase1_data_collection}

We construct the news corpus via a three-stage pipeline (Fig.~\ref{fig:phase1_flow}):
(i) base corpus sampling, (ii) value-driven supplementation for coverage and balance, and
(iii) unified filtering and cleaning for consistency and safety.
We draw articles primarily from Uknow~\cite{gong2024uknow} and English Wikinews (2022--2025) via the MediaWiki Action API\footnote{\url{https://en.wikinews.org/w/api.php}.}.
Full procedural details, including source-domain lists and filtering criteria, are provided in Appendix~\ref{app:phase1_data_collection_sources}.

\subsection{Phase-2: Event Knowledge Base Construction}
\label{sec:phase2_event_kb}

\begin{figure}[t]
    \centering
    \includegraphics[width=\textwidth, trim=0mm 30mm 40mm 22mm, clip]{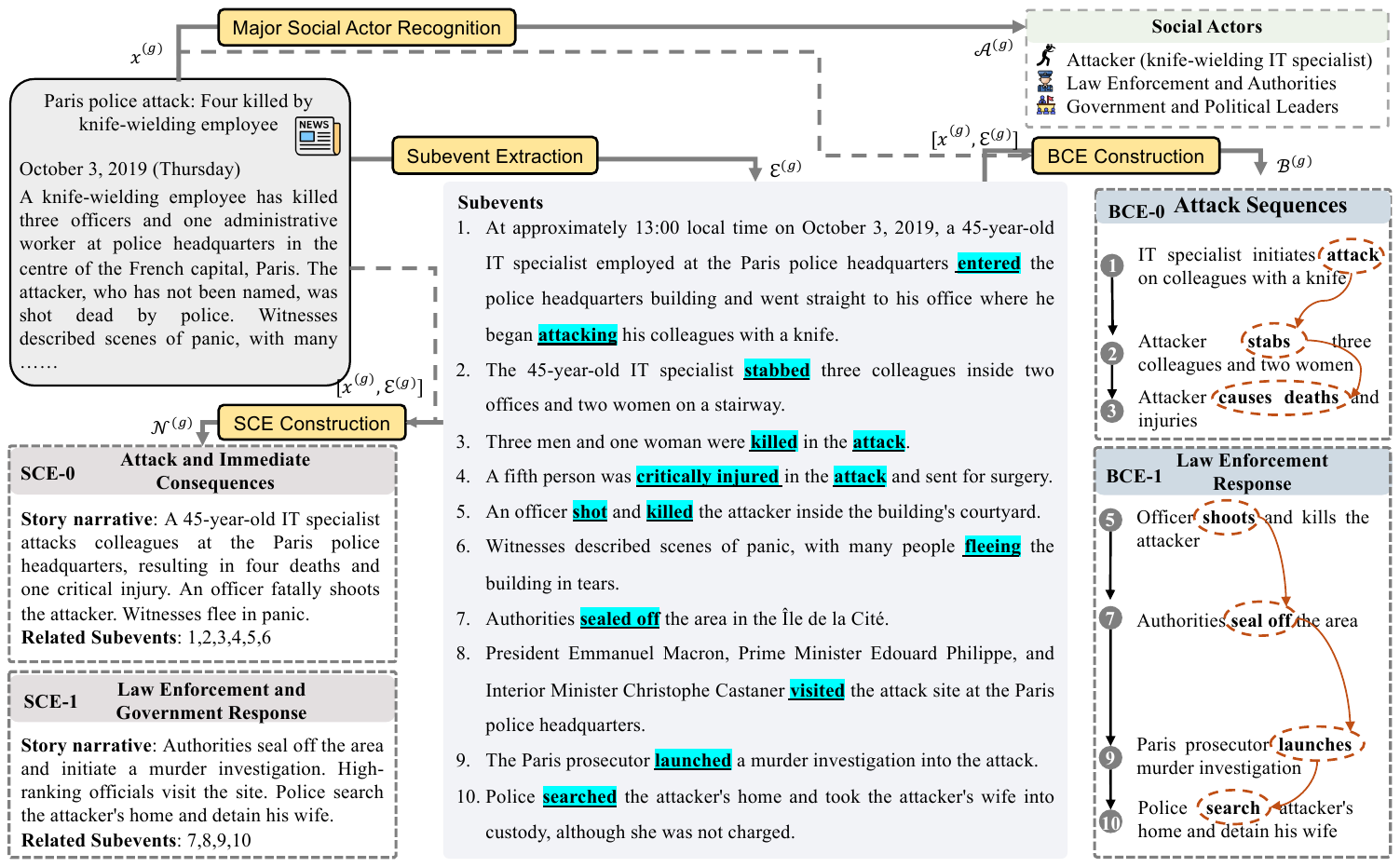}
    \caption{An illustrative example of the \emph{joint} construction of the event-centric knowledge base.
    }
    \label{fig:phase2_flow}
\end{figure}

To support fine-grained human value understanding, we construct an event-centric knowledge base from the multi-source news corpus described in Phase-1.
Within each article, we build a multi-level semantic hierarchy consisting of sentence-grounded subevents, two complementary types of composite event units (BCEs and SCEs), and a set of major social actors (Fig.~\ref{fig:phase2_flow}).
This hierarchy is designed to make value inference \emph{localizable} through evidence-grounded subevents and more \emph{diagnosable} through contextual aggregation across semantic scopes.
Detailed schema definitions and construction rules are provided in Appendix~\ref{app:event_kb_details}.

\paragraph{Subevents.}
To move beyond coarse article-level representations, we extract subevents as fine-grained local event units, following prior benchmark designs that operationalize events at sentence-level granularity for long-context and temporally complex event analysis \cite{zhang2024analyzing}.
Each subevent is associated with a short natural-language summary and supporting evidence sentences from the source article, enabling traceable value annotation.
For subevent characterization, we use the WikiEvents Ontology~\cite{li2021document} as a conceptual reference.

\paragraph{Composite events.}
While subevents provide localized evidence, value cues in news are often distributed across related actions, responses, and consequences.
To capture such context beyond isolated subevents, we construct two complementary composite event units, following prior work on event composition, relational inference, and event-centric document modeling \cite{chakravarthy1994composite,bosselut2019comet,hwang2021comet,li2021future,keith2023survey}.
A \textbf{BCE} groups closely linked subevents into a localized behavior or response chain, supporting value attribution over procedural continuity and immediate outcomes.
A \textbf{SCE} groups more broadly related subevents into a larger story segment, supporting value attribution over wider narrative development and downstream consequences.
We distinguish these two views because value judgments in news may depend either on a tightly connected local process or on a broader narrative segment that integrates follow-up developments.
In Fig.~\ref{fig:phase2_flow}, for instance, BCEs organize linked attack and response subevents into local chains, whereas SCEs capture broader story segments such as the attack episode and the institutional response storyline.

\paragraph{Major social actors.}
For each article, we identify major social actors involved in the described processes, including individuals, institutions, governments, companies, and social groups. 
To reduce ambiguity from aliases, coreference, and varying mention specificity, we canonicalize actor mentions and, when appropriate, abstract them into role-based actors defined by their event participation. 
This actor layer supports consistent attribution of value judgments across units and levels.

\begin{figure}[t]
    \centering
    \includegraphics[width=\textwidth, trim=5mm 45mm 5mm 15mm, clip]{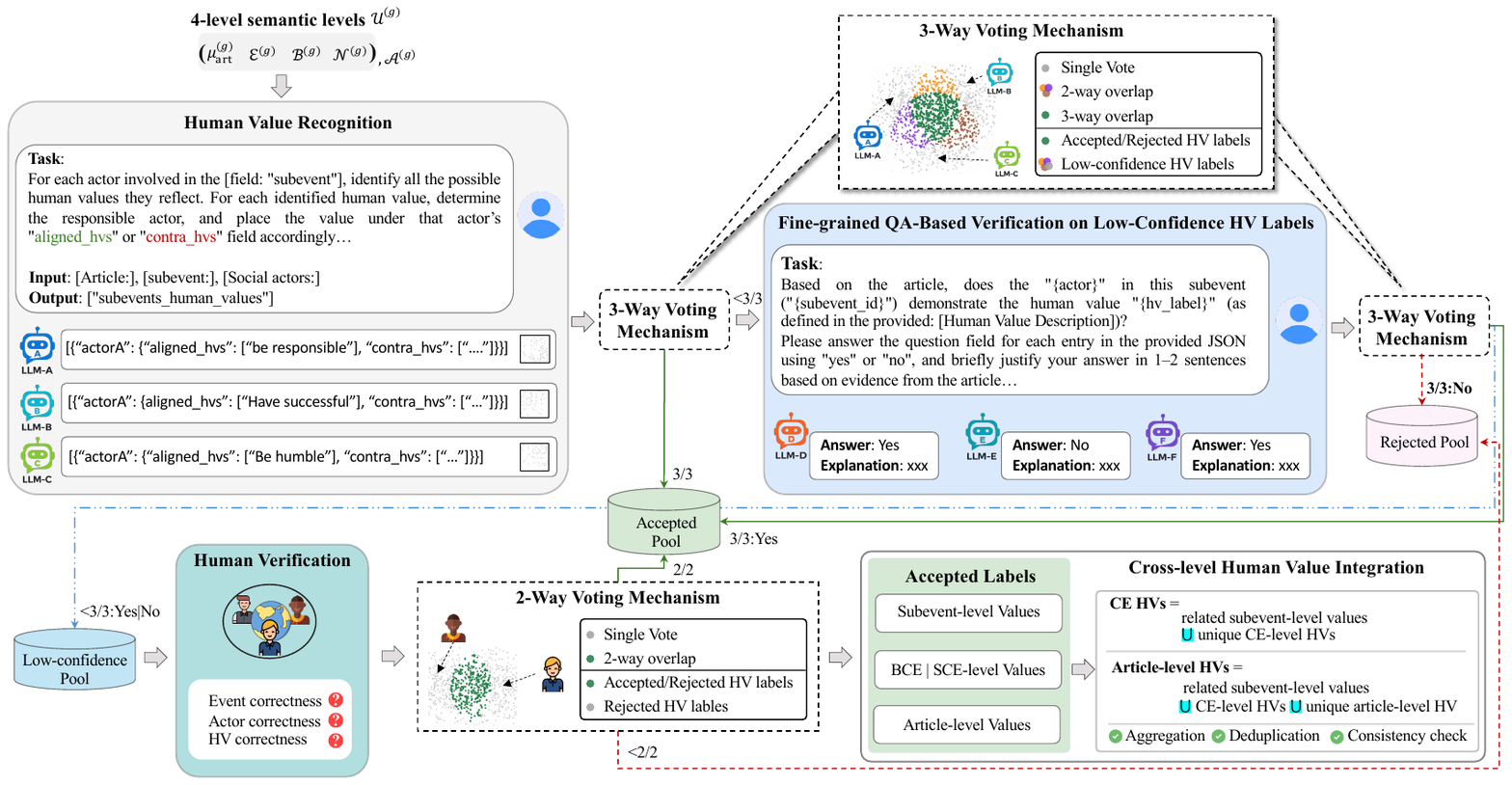}
    \caption{Phase-3 human value annotation pipeline for NEVU.
    }
    \label{fig:phase3_hv_annotation}
\end{figure}

\subsection{Phase-3: Human Value Annotation Pipeline}
\label{sec:phase3_hv_annotation}

As illustrated in Fig.~\ref{fig:phase3_hv_annotation}, we adopt an \emph{LLM-assisted, staged} annotation pipeline to construct actor-conditioned human value annotations. 
The pipeline is designed to mitigate single-model bias, enforce evidence grounding, and reduce human annotation cost through progressive verification. 
Values are annotated over \mbox{(unit, actor)} pairs across the Article, Subevent, BCE, and SCE levels, allowing comparison across local event units and broader composite contexts.
Detailed annotation procedures and verification rules are provided in Appendix~\ref{app:phase3_details}.

\paragraph{Stage 1: LLM-based Candidate Identification}
\label{sec:stage1_candidate_identification}

We first use multiple LLMs to propose candidate directed value labels for each $(u,a)$ pair.
To improve precision and reduce single-model bias, only candidates with strong cross-model agreement are retained, while uncertain cases are deferred to the next verification stage.

\paragraph{Stage 2: QA-based verification of low-confidence labels.}
Low-confidence candidates from Stage 1 are passed to a secondary verification stage.
Rather than generating labels from scratch, this stage evaluates whether a candidate value-direction judgment is supported by the available evidence for the target actor and unit.
Cases that remain unresolved after model verification are sent to human verification.

\begin{wrapfigure}[14]{r}[0pt]{0.5\textwidth}
  \centering
  \vspace{-45pt}
  \includegraphics[width=\linewidth, trim=5mm 100mm 120mm 5mm, clip]{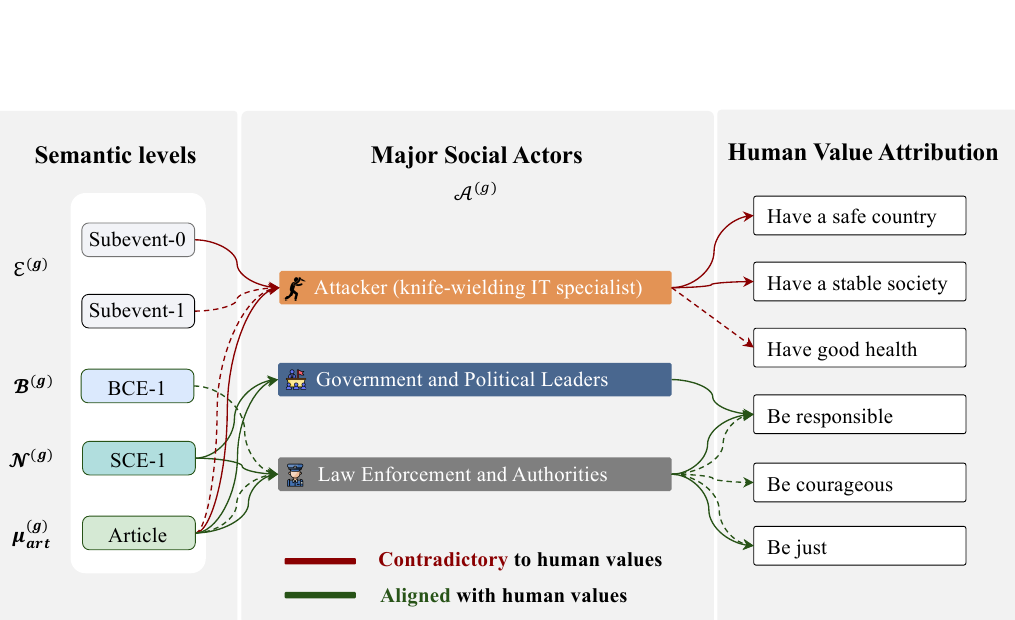}
  \caption{\small Illustrative examples of actor-conditioned human value annotations across article, subevent, BCE, and SCE levels.}
  \label{fig:phase3_hv_examples}
  \vspace{30pt}
\end{wrapfigure}

\paragraph{Stage 3: Human verification.}
For low-confidence candidates from Stage 2, we conduct targeted human verification to improve annotation reliability.
This step focuses on ambiguous cases rather than exhaustive manual review of all released value instances.

\paragraph{Stage 4: Cross-level value integration.}
Finally, we integrate verified labels across semantic levels from fine to coarse granularity. 
Composite-event units (BCEs and SCEs) aggregate supported value evidence from linked subevents, while also allowing additional labels that become clearer only when related subevents are viewed together as a localized chain or a broader narrative segment. 
We then normalize the resulting labels through aggregation, deduplication, and cross-level consistency checks to produce the final gold actor-conditioned directed value annotations.

\paragraph{Finalization breakdown of released labels.}
For transparency, we report how released gold value instances are finalized across the annotation pipeline.
Of the 168{,}061 released value instances, 77{,}913 (46.36\%) were retained in \textbf{Stage~1} via strong cross-model agreement during LLM-based candidate identification, 75{,}048 (44.66\%) were finalized in \textbf{Stage~2} via QA-based verification, and 15{,}100 (8.98\%) were finalized in \textbf{Stage~3} via targeted human verification.
\textbf{Stage~4} performs cross-level integration and normalization after verification, and is therefore not treated as a separate label-finalization route in this summary.
These statistics clarify that NEVU is constructed through a staged annotation and verification pipeline rather than a single-pass LLM labeling procedure.

\begin{table}[t]
\centering
\footnotesize
\setlength{\tabcolsep}{3.0pt}
\renewcommand{\arraystretch}{1.05}

\caption{Split-wise scale and annotation statistics of actor-conditioned human value annotations.
Align.\% is $100\times \text{Align.}/(\text{Align.}+\text{Contra.})$.}
\label{tab:scale_structure_all}

\begin{adjustbox}{max width=\columnwidth}
\begin{tabular}{llrrrrrr rrrrr}
\toprule
\multicolumn{2}{c}{\textbf{Subset}} &
\multicolumn{4}{c}{\textbf{Units}} &
\multicolumn{2}{c}{\textbf{Actors}} &
\multicolumn{5}{c}{\textbf{Value annotations}} \\
\cmidrule(lr){1-2}\cmidrule(lr){3-6}\cmidrule(lr){7-8}\cmidrule(lr){9-13}

\textbf{Grp} & \textbf{Split} &
\textbf{Art.} & \textbf{Sub.} & \textbf{BCE} & \textbf{SCE} &
\textbf{Act.} & Avg. &
\textbf{Inst.} & \textbf{Align.} & \textbf{Contra.} & \textbf{Aligned ratio} & \textbf{(u, a)} \\
\midrule
Total & dev   & 286  & 2501  & 951  & 799  & 1120 & 3.993 & 16354  & 12707 & 3647  & 77.7 & 7049  \\
Total & test  & 574  & 5430  & 1872 & 1619 & 2317 & 4.233 & 34079  & 25599 & 8480  & 75.1 & 14875 \\
Total & train & 2005 & 18395 & 6575 & 5582 & 7679 & 4.189 & 117628 & 90474 & 27154 & 76.9 & 50981 \\
\addlinespace[2pt]
Sub   & dev   & 283  & 2476  & 944  & 792  & 1109 & 3.996 & 7907   & 6059  & 1848  & 76.6 & 2500  \\
Sub   & test  & 568  & 5388  & 1855 & 1601 & 2295 & 4.236 & 15869  & 11920 & 3949  & 75.1 & 5000  \\
Sub   & train & 1991 & 18311 & 6536 & 5549 & 7632 & 4.191 & 62110  & 47457 & 14653 & 76.4 & 20000 \\
\bottomrule
\end{tabular}
\end{adjustbox}

\vspace{2pt}
\raggedright
\footnotesize
\noindent \textit{Notes.}
Act.=unique actors; Avg.=mean actors per article; (u,a)=annotated unit--actor pairs; Inst.=directed value instances (Inst.=Align.+Contra.).
\end{table}

\subsection{Dataset Statistics and Visualizations}
\label{sec:dataset_statistics}

Table~\ref{tab:scale_structure_all} reports split-wise scale and annotation statistics of NEVU, together with the sampled subset (\textbf{Sub}) used in downstream experiments.
The full benchmark contains 2,005/286/574 articles in train/dev/test, which expand into 18,395/2,501/5,430 subevents, 6,575/951/1,872 BCEs, and 5,582/799/1,619 SCEs, respectively.
NEVU is actor-centric: across the three splits, we annotate 7,679/1,120/2,317 unique abstract social actors, with an average of about 4.0--4.2 actors per article.
At the annotation level, the benchmark contains 117,628/16,354/34,079 directed value instances in train/dev/test, defined over annotated unit--actor pairs.
Across splits, aligned instances account for about three quarters of all annotations (75.1\%--77.7\%), while contradictory instances remain substantial, providing meaningful coverage for direction-aware evaluation.
The \textbf{Sub} rows correspond to sampled subsets drawn from the fixed train/dev/test partitions and are used only to reduce experimental cost in selected analyses and ablations, without altering the benchmark split protocol.

\begin{figure}[t]
    \centering
    \includegraphics[width=\textwidth, clip]{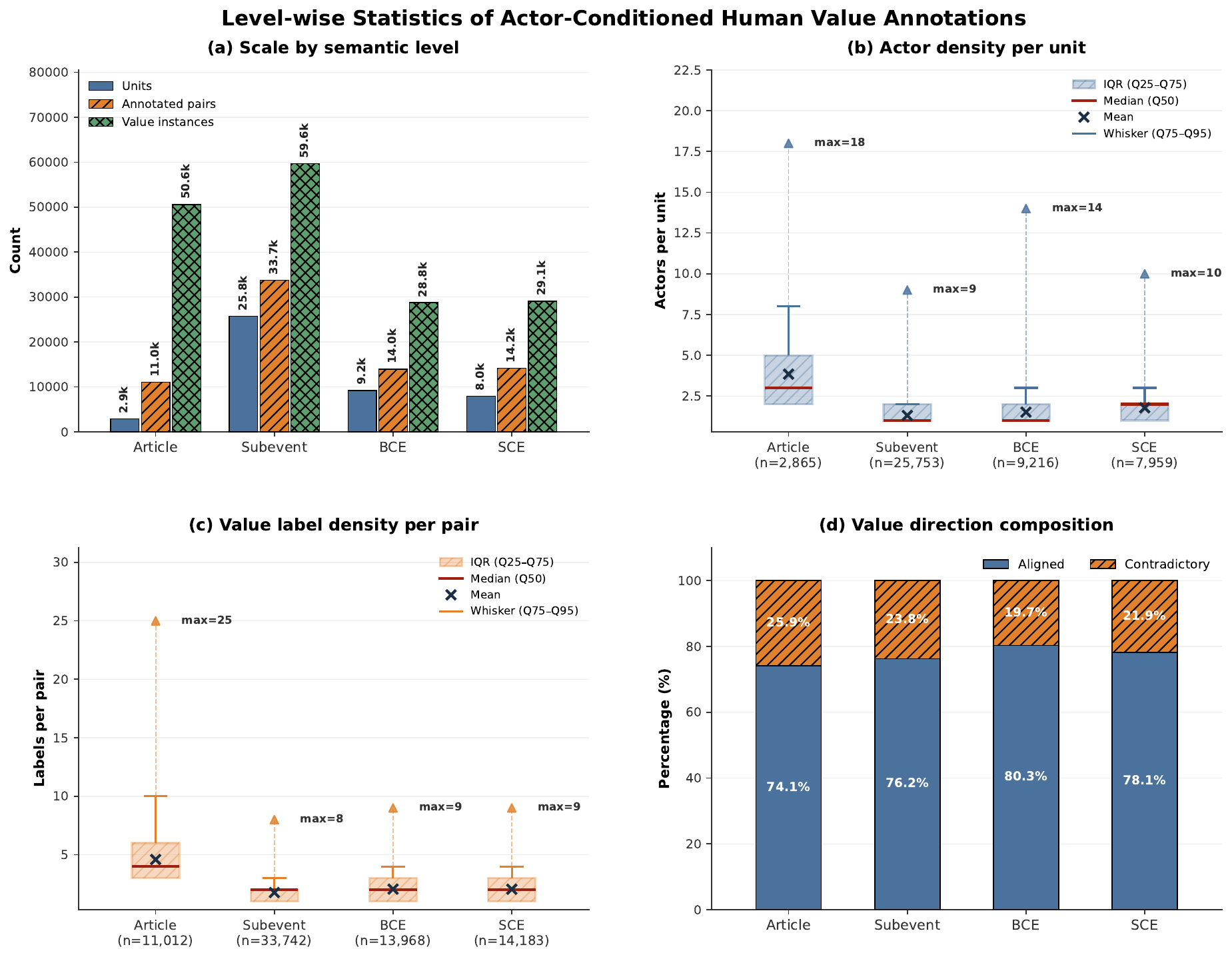}
    \caption{Level-wise annotation statistics on the full dataset ($s=\texttt{all}$; train+dev+test): 
(a) numbers of units, annotated unit--actor pairs, and directed value instances; 
(b) actor density per unit; 
(c) directed value labels per unit--actor pair; 
(d) aligned vs.\ contradictory label ratios across semantic levels.}
    \label{fig:levelwise_annotation_statistics_4panel}
\end{figure}

Figure~\ref{fig:levelwise_annotation_statistics_4panel} summarizes level-wise annotation characteristics on the full dataset ($s=\texttt{all}$; train+dev+test).
Subevents dominate the dataset scale, accounting for the largest number of units, annotated unit--actor pairs, and directed value instances, whereas BCEs and SCEs provide intermediate-scale composite structures between localized subevents and full articles.
Article units contain the richest actor context, while subevents are much more localized and usually center on a single actor.
At the same time, article-level unit--actor pairs carry the densest value annotations overall, and composite units (BCE/SCE) remain denser than subevents in directed value labels per pair.
Together, these patterns suggest that subevents provide stronger locality and cleaner actor focus, whereas composite units more often concentrate multiple value-relevant actions and consequences within the same semantic structure.
Aligned labels dominate across all levels, with BCE showing the highest aligned ratio.

\subsection{Dataset Quality Assessment}
\label{sec:quality_assessment}

We conduct a manual quality assessment of NEVU on a stratified sample across the four semantic granularities and report Soft/Hard scores. 
Overall, the results indicate strong reliability for actor abstraction and event construction, with N-weighted Actor Concept Validity, Actor Grounding Adequacy, and Human Value Correctness reaching 94.8/93.4, 92.1/90.3, and 84.6/83.2, respectively. 
Detailed evaluation settings, dimension definitions, and full results are provided in Appendix~\ref{app:human_evaluation}.

\section{Benchmark Task and Evaluation Protocol}
\label{sec:method}

Building on the problem formulation in Section~\ref{sec:problem}, we instantiate Human Value Recognition in NEVU as a directed multi-label prediction task over annotated unit--actor pairs and evaluate it under a unified benchmark protocol.

\subsection{Benchmark Instantiation}
\label{sec:task_hvr}

Evaluation in NEVU is performed over annotated unit--actor pairs $(u,a)$ derived from articles $g$, where the target unit belongs to one of four semantic levels:
\begin{equation}
\label{eq:unit-level}
\ell(u)\in\{\texttt{article},\texttt{subevent},\texttt{BCE},\texttt{SCE}\}.
\end{equation}
For each instance, the model outputs two direction-specific sets of Level-1 value labels, corresponding to aligned ($+$) and contradictory ($-$) predictions:
\begin{equation}
\widehat{\Phi}^{(1)}(u,a)=\big(\widehat{\Phi}^{(1)+}(u,a),\ \widehat{\Phi}^{(1)-}(u,a)\big),
\qquad
\widehat{\Phi}^{(1)\pm}(u,a)\subseteq \mathcal{V}^{(1)}.
\end{equation}

\paragraph{Input representation.}
For each unit--actor pair $(u,a)$, the model input consists of the unit text, grounded evidence, contextual information, the unit level, and the target actor:
\begin{equation}
\label{eq:model-input}
\mathbf{I}(u,a)=\big(\texttt{unit\_text}(u),\ \texttt{evidence}(u),\ \texttt{context}(u),\ \texttt{unit\_level}=\ell(u),\ \texttt{actor}=a\big).
\end{equation}
At the article level, \texttt{unit\_text} is the full article.
At the subevent level, it is the subevent summary, with grounded sentences as \texttt{evidence} and neighboring sentences as \texttt{context}.
For BCE and SCE, \texttt{unit\_text} combines the composite description with linked subevent summaries, while \texttt{evidence} and \texttt{context} aggregate the corresponding grounded sentences and local context.

\paragraph{Label space and mapping.}
NEVU supports evaluation at both Level-1 (54 labels) and Level-2 (20 labels).
Models predict only Level-1 labels, which are then deterministically mapped to Level-2 using a fixed function
$f:\mathcal{V}^{(1)}\rightarrow \mathcal{V}^{(2)}$ while preserving direction:
\begin{equation}
\widehat{\Phi}^{(2)\pm}(u,a)=\{\, f(v)\mid v\in\widehat{\Phi}^{(1)\pm}(u,a)\,\}.
\end{equation}
We report results at both levels under the same prediction process.

\subsubsection{Long-context evaluation protocol (map--reduce)}
\label{sec:map-reduce}

When \texttt{unit\_text} exceeds the model context budget, we apply a map--reduce inference procedure \cite{zhou2025llm}.
Only \texttt{unit\_text} is split into $K$ fixed-length chunks, while \texttt{evidence}, \texttt{context}, \texttt{unit\_level}, and the target actor remain unchanged.
Predictions are made for each chunk and merged by direction-specific union with de-duplication:
\begin{equation}
\widehat{\Phi}^{(1)\pm}(u,a)=\bigcup_{k=1}^{K}\widehat{\Phi}^{(1)\pm}_k(u,a).
\end{equation}

\subsection{Evaluation Metrics}
\label{sec:eval_metrics}

For evaluation, both gold and predicted outputs for each unit--actor pair $(u,a)$ are represented as sets of directed value labels, where each label is a pair $(v,d)$ with $v\in\mathcal{V}^{(1)}$ and $d\in\{+,-\}$.
We denote the gold set by $Y(u,a)$ and the predicted set by $\widehat{Y}(u,a)$, and compare them over all instances in the benchmark.

\paragraph{Micro-F1.}
We report Micro-F1 by aggregating true positives, false positives, and false negatives over all evaluated $(u,a)$ instances:
\begin{equation}
\mathrm{Micro\text{-}P}=\frac{\mathrm{TP}}{\mathrm{TP}+\mathrm{FP}},\quad
\mathrm{Micro\text{-}R}=\frac{\mathrm{TP}}{\mathrm{TP}+\mathrm{FN}},\quad
\mathrm{Micro\text{-}F1}=\frac{2\cdot \mathrm{Micro\text{-}P}\cdot \mathrm{Micro\text{-}R}}
{\mathrm{Micro\text{-}P}+\mathrm{Micro\text{-}R}}.
\end{equation}

\paragraph{Macro-F1 (gold-supported).}
We also report Macro-F1 over the directed label space by averaging per-label F1 scores.
To avoid distortion from labels absent in the evaluated split, we compute this metric in a gold-supported manner, averaging only over labels that appear at least once in the gold annotations.

\paragraph{Direction Reverse Rate (DRR).}
To quantify direction errors, we report Direction Reverse Rate (DRR), the rate at which a model predicts the correct value identity but assigns the opposite direction.
For any $(u,a)$ where the same value appears with both directions in the gold annotation, that value is excluded from the DRR denominator because its direction is ambiguous.
Lower DRR indicates fewer aligned/contradictory reversals.

\begin{table}[t]
\centering
\small
\setlength{\tabcolsep}{6pt}
\caption{Model groups used in the Human Value Recognition benchmark.
G1 and G2 are evaluated in a prompting-only setting, while G3 applies LoRA fine-tuning to the same open-weight backbones.
All groups use a shared prompt template and actor-centric JSON output schema.}
\begin{tabular}{p{0.28\linewidth} p{0.68\linewidth}}
\toprule
\textbf{Group} & \textbf{Models} \\
\midrule
G1: Proprietary API
& OpenAI gpt-5.2 \cite{openai_gpt52_model_2025}, gpt-4.1 \cite{openai_gpt4_2024}, gpt-o3 \cite{openai_o3_o4mini_release_2025}; Claude Haiku \cite{anthropic_haiku45_system_card_2025}/Sonnet \cite{anthropic_sonnet45_system_card_2025}/Opus 4.5 \cite{anthropic_opus45_system_card_2025}; Gemini-2.5-Pro \cite{gemini25_2025}; Deepseek-chat \cite{deepseekv32_2025}, Deepseek-reasoner \cite{deepseek_reasoner_api_2026} \\
\midrule
G2: Open-weight (w/o fine-tuning)
& Llama-3.1-8B-Instruct \cite{weerawardhena2025llama31foundationaisecurityllm8binstructtechnicalreport}, Qwen3-8B \cite{yang2025qwen3technicalreport}, Ministral-3-8B-Instruct-2512 \cite{liu2026ministral3}, Phi-3.5-mini-instruct \cite{abdin2024phi3technicalreporthighly} \\
\midrule
G3: Open-weight (fine-tuned)
& Llama-3.1-8B-Instruct + LoRA, Qwen3-8B + LoRA, Ministral-3-8B-Instruct-2512 + LoRA, Phi-3.5-mini-instruct + LoRA \\
\bottomrule
\end{tabular}
\label{tab:baseline-hvr}
\end{table}

\section{Experiments}
\subsection{Experimental Settings}
\label{sec:exp_settings}

We benchmark a diverse set of representative LLMs on the Human Value Recognition task under a unified protocol.
The evaluated models are grouped into three categories (Table~\ref{tab:baseline-hvr}) based on access mode and whether task-specific fine-tuning is applied.
Performance is measured using Micro-F1, Macro-F1, and DRR, as defined in Sec.~\ref{sec:eval_metrics}.

\paragraph{Unified prompting protocol.}
To ensure fair comparison across proprietary and open-weight models, all models are evaluated with a shared prompt template and actor-centric JSON output schema.
Given a textual unit at one of the four semantic levels and a target actor, models predict directed Level-1 value labels.
Level-2 labels are obtained deterministically by mapping the predicted Level-1 labels to the coarser taxonomy.

\paragraph{G1: Proprietary API models (prompted, no task-specific training).}
G1 consists of proprietary LLMs accessed via APIs.
They are evaluated in a prompting-only setting and serve as high-performance reference systems under the shared protocol.

\paragraph{G2: Open-weight models (prompted, w/o fine-tuning).}
G2 includes open-weight instruction-tuned backbones evaluated in the same prompting-only setting.
This group provides fully reproducible baselines for actor-conditioned value recognition without task-specific adaptation.

\paragraph{G3: Open-weight models with LoRA fine-tuning.}
To assess the effect of lightweight task adaptation, we fine-tune the same backbones as in G2 with LoRA on the training split and evaluate them under the same protocol.
This group measures the gains from supervised adaptation while keeping the model family and output schema fixed.

\subsection{Human Value Recognition Results and Discussion}
\label{sec:hvr-results-discussion}

\begin{table}[t]
\centering
\small
\setlength{\tabcolsep}{2.4pt}
\renewcommand{\arraystretch}{1.08}
\caption{Human Value Recognition results on 54 fine-grained Level-1 human values.
G1: proprietary API LLMs (prompting only); G2: open-weight LLMs (prompting only); G3: open-weight LLMs with LoRA fine-tuning.
Best-in-group is underlined; overall best is underlined in bold.}
\resizebox{\linewidth}{!}{
\begin{tabular}{llccccc ccccc ccccc}
\toprule
\multirow{2}{*}{\textbf{Group}} & \multirow{2}{*}{\textbf{Model}}
& \multicolumn{5}{c}{\textbf{Micro-F1}} 
& \multicolumn{5}{c}{\textbf{Macro-F1}} 
& \multicolumn{5}{c}{\textbf{DRR$\downarrow$}} \\
\cmidrule(lr){3-7}\cmidrule(lr){8-12}\cmidrule(lr){13-17}
& 
& \textbf{Art.} & \textbf{Sub.} & \textbf{BCE} & \textbf{SCE} & \textbf{Overall}
& \textbf{Art.} & \textbf{Sub.} & \textbf{BCE} & \textbf{SCE} & \textbf{Overall}
& \textbf{Art.} & \textbf{Sub.} & \textbf{BCE} & \textbf{SCE} & \textbf{Overall} \\
\midrule

\multirow{9}{*}{G1}
& Claude Haiku 4.5
& 43.05 & 41.02 & 38.38 & 39.69 & 40.87
& 28.87 & 33.13 & 28.18 & 32.93 & 30.11
& 3.44 & 1.72 & 2.41 & 2.43 & 2.74 \\
& Claude Sonnet 4.5
& 46.37 & 44.91 & 42.17 & 43.22 & 44.43
& 32.54 & 39.03 & 35.41 & 36.37 & 34.73
& 2.61 & 1.22 & 1.97 & 2.19 & 2.15 \\
& Claude Opus 4.5
& 47.76 & \underline{46.68} & \underline{44.47} & \underline{45.68} & 46.37
& 34.86 & 41.00 & 37.07 & 37.33 & 37.25
& 2.82 & \underline{1.00} & 2.00 & 1.92 & 2.16 \\
& Deepseek-chat
& 36.62 & 39.76 & 39.04 & 40.98 & 38.67
& 23.05 & 30.32 & 29.95 & 30.23 & 26.82
& 2.64 & 1.36 & 2.31 & 2.53 & 2.32 \\
& Deepseek-reasoner
& 43.47 & 43.45 & 43.87 & 45.23 & 43.94
& 31.76 & 36.81 & 34.06 & 37.73 & 34.92
& 1.65 & 1.40 & 1.42 & 1.16 & 1.47 \\
& OpenAI gpt-4.1
& 42.71 & 41.67 & 40.55 & 41.63 & 41.82
& 29.44 & 34.31 & 32.36 & 32.89 & 31.12
& 2.95 & 1.22 & 1.97 & 1.92 & 2.25 \\
& OpenAI gpt-5.2
& 42.48 & 44.97 & 44.38 & 45.31 & 43.96
& 28.77 & 34.52 & 34.09 & 35.33 & 32.00
& 2.10 & 1.22 & 1.80 & 1.85 & 1.83 \\
& OpenAI o3
& 37.98 & 42.62 & 41.48 & 42.60 & 40.56
& 27.77 & 33.69 & 32.59 & 32.77 & 31.22
& \underline{1.09} & 1.18 & \underline{1.09} & \underline{1.03} & \underline{1.09} \\
& Gemini 2.5 Pro
& \underline{53.36} & 45.49 & 42.66 & 43.04 & \underline{47.05}
& \underline{42.71} & \underline{42.23} & \underline{37.24} & \underline{37.90} & \underline{39.87}
& 3.10 & 1.50 & 2.10 & 1.88 & 2.39 \\
\midrule

\multirow{4}{*}{G2}
& Ministral-3-8B
& \underline{23.67} & \underline{26.00} & \underline{26.27} & \underline{27.31} & \underline{25.49}
& \underline{14.18} & \underline{20.21} & \underline{17.99} & \underline{19.61} & \underline{17.17}
& \underline{0.04} & 0.03 & 0.03 & 0.03 & 0.04 \\
& Phi-3.5-mini
& 13.31 & 7.17 & 9.04 & 10.02 & 10.56
& 6.64 & 4.57 & 4.90 & 6.76 & 5.73
& \underline{0.04} & \underline{0.02} & \underline{0.02} & \underline{0.02} & \underline{0.03} \\
& Qwen3-8B
& 18.83 & 13.67 & 14.88 & 14.88 & 16.06
& 9.69 & 8.65 & 9.11 & 10.48 & 9.35
& 3.59 & 1.14 & 1.83 & 2.05 & 2.52 \\
& Llama-3.1-8B
& 15.19 & 11.79 & 11.29 & 11.40 & 12.97
& 6.02 & 6.39 & 5.06 & 6.46 & 5.94
& 7.81 & 3.36 & 4.24 & 4.35 & 5.66 \\
\midrule

\multirow{4}{*}{G3}
& Ministral-3-8B
& 57.35 & 50.04 & 49.58 & 52.52 & 53.71
& 42.75 & 42.07 & 36.66 & 39.85 & 41.73
& 0.04 & \underline{\textbf{0.01}} & \underline{\textbf{0.01}} & \underline{\textbf{0.01}} & \underline{\textbf{0.02}} \\
& Phi-3.5-mini
& 54.35 & 48.40 & 51.05 & 53.32 & 52.53
& 37.91 & 38.75 & 36.44 & 40.20 & 38.90
& 0.05 & \underline{\textbf{0.01}} & \underline{\textbf{0.01}} & \underline{\textbf{0.01}} & 0.03 \\
& Qwen3-8B
& 56.77 & 50.86 & \underline{\textbf{53.15}} & 53.41 & 54.44
& 41.21 & 41.07 & \underline{\textbf{42.72}} & \underline{\textbf{43.42}} & 42.42
& \underline{\textbf{0.03}} & \underline{\textbf{0.01}} & \underline{\textbf{0.01}} & \underline{\textbf{0.01}} & \underline{\textbf{0.02}} \\
& Llama-3.1-8B
& \underline{\textbf{57.77}} & \underline{\textbf{53.50}} & 52.30 & \underline{\textbf{53.70}} & \underline{\textbf{55.26}}
& \underline{\textbf{43.02}} & \underline{\textbf{45.05}} & 40.89 & 42.44 & \underline{\textbf{44.29}}
& 3.10 & 0.75 & 0.61 & 0.92 & 1.78 \\
\bottomrule
\end{tabular}
}
\label{tab:hvr-l1-merged-levels}
\end{table}

Table~\ref{tab:hvr-l1-merged-levels} reports Level-1 results over the 54 fine-grained values across the four semantic levels and Overall.
Appendix Table~\ref{app:hvr-l2-merged-levels} provides the corresponding Level-2 view obtained by deterministic mapping from Level-1 predictions.

\paragraph{Overall performance and group comparison.}
On Level-1, \textbf{Gemini 2.5 Pro} is the strongest model in \textbf{G1}, achieving \textbf{47.05} Overall Micro-F1 and \textbf{39.87} Overall Macro-F1.
However, the strongest \textbf{G3} models outperform all prompting-only systems overall: \textbf{Llama-3.1-8B+LoRA} reaches \textbf{55.26}/\textbf{44.29} (Micro/Macro), followed by \textbf{Qwen3-8B+LoRA} with \textbf{54.44}/\textbf{42.42}.
This shows that lightweight supervised adaptation substantially improves benchmark performance beyond prompting alone.

\paragraph{Effect of task adaptation: G2 $\rightarrow$ G3.}
Across all open-weight backbones, LoRA training yields large gains on Level-1.
Overall Micro-F1 increases from \textbf{25.49} to \textbf{53.71} for Ministral-3-8B, from \textbf{16.06} to \textbf{54.44} for Qwen3-8B, from \textbf{10.56} to \textbf{52.53} for Phi-3.5-mini, and from \textbf{12.97} to \textbf{55.26} for Llama-3.1-8B.
These gains indicate that NEVU provides a strong supervised learning signal, and that prompting-only performance substantially underestimates what open-weight backbones can achieve after lightweight adaptation.

\paragraph{Micro, Macro, and direction sensitivity.}
Across all groups, Macro-F1 is consistently lower than Micro-F1 under the 54-label taxonomy, indicating that models are more reliable on frequent values than on rarer ones.
Task adaptation narrows this gap to some extent, as seen in the stronger Macro-F1 of \textbf{G3} models relative to their prompting-only counterparts.
DRR provides a complementary view by isolating direction reversals.
Most \textbf{G3} models achieve extremely low DRR on the composite levels, often around \textbf{0.01}, showing that fine-tuning strongly reduces aligned versus contradictory reversals.
At the same time, the best F1 model does not always achieve the lowest DRR: \textbf{Llama-3.1-8B+LoRA} has the highest Overall F1 but an Overall DRR of \textbf{1.78}, whereas \textbf{OpenAI o3} achieves the lowest Overall DRR in \textbf{G1} at \textbf{1.09} without reaching the top F1 tier.

\paragraph{Robustness across semantic levels.}
The level-wise results show that performance patterns differ across local and composite units.
Within \textbf{G1}, \textbf{Gemini 2.5 Pro} is strongest at the article level (\textbf{53.36} Micro-F1) and remains competitive on BCE and SCE.
However, the strongest \textbf{G3} models are more consistent across levels.
For example, \textbf{Qwen3-8B+LoRA} achieves \textbf{53.15} on BCE and \textbf{53.41} on SCE, both close to its article-level score, indicating relatively stable recognition under hierarchical event representations.

\paragraph{Level-2 results.}
The same ranking pattern largely holds at Level-2: \textbf{Gemini 2.5 Pro} remains the strongest model in \textbf{G1}, while \textbf{G3} achieves the best overall results.
Because Level-2 is deterministically mapped from Level-1 outputs, the higher scores mainly reflect the reduction of fine-grained confusions under the coarser taxonomy.

\begin{table}[t]
\centering
\small
\setlength{\tabcolsep}{5pt}
\renewcommand{\arraystretch}{1.15}
\caption{Subset statistics (Level-1) for the test-sampled data used in our two diagnostic analyses: hierarchical consistency profiling (HCP) and controlled grouping test (CGT).}
\begin{tabular}{llrrrrrr}
\toprule
\textbf{Set.} &
\textbf{Lvl.} &
\textbf{Units} &
\textbf{(u,a)} &
\textbf{Inst.} &
\textbf{Align.} &
\textbf{Contra.} &
\textbf{Aligned ratio} \\
\midrule
HCP & Art. & 192  & 3513 & 753  & 2634 & 879 & 75.0 \\
HCP & BCE  & 650  & 1991 & 955  & 1596 & 395 & 80.2 \\
HCP & SCE  & 569  & 1979 & 967  & 1585 & 394 & 80.1 \\
HCP & Sub. & 1878 & 4152 & 2347 & 3156 & 996 & 76.0 \\
\midrule
CGT & BCE  & 412  & 509  & 1078 & 854  & 224 & 79.2 \\
CGT & SCE  & 353  & 462  & 905  & 718  & 187 & 79.3 \\
\bottomrule
\end{tabular}
\label{tab:sampled_test_hv_stats}
\end{table}

\subsection{Hierarchical Consistency Profiling}
\label{sec:hier_consistency}

\begin{table}[t]
\centering
\scriptsize
\setlength{\tabcolsep}{3.0pt}
\renewcommand{\arraystretch}{1.12}
\caption{Hierarchical Consistency Profiling on Level-1 (54 values).
Left: task scores; Right: cross-level consistency scores.
$\Delta$Overall denotes the pp gap to the gold Overall consistency.}
\resizebox{\textwidth}{!}{%
\begin{tabular}{l ccc ccc ccc ccc ccc | ccc ccc ccc ccc | c}
\toprule
& \multicolumn{15}{c|}{Task (\%): Micro-F1 / Micro-P / Micro-R}
& \multicolumn{12}{c|}{Consistency (\%): mean-F1 / mean-P / mean-R}
& \\
\cmidrule(lr){2-16}\cmidrule(lr){17-28}
Model
& \multicolumn{3}{c}{Art.} & \multicolumn{3}{c}{Sub.} & \multicolumn{3}{c}{BCE} & \multicolumn{3}{c}{SCE} & \multicolumn{3}{c|}{Overall}
& \multicolumn{3}{c}{Art.} & \multicolumn{3}{c}{BCE} & \multicolumn{3}{c}{SCE} & \multicolumn{3}{c|}{Overall}
& $\Delta$Overall (to Gold) \\
\cmidrule(lr){2-4}\cmidrule(lr){5-7}\cmidrule(lr){8-10}\cmidrule(lr){11-13}\cmidrule(lr){14-16}
\cmidrule(lr){17-19}\cmidrule(lr){20-22}\cmidrule(lr){23-25}\cmidrule(lr){26-28}
& F1 & P & R & F1 & P & R & F1 & P & R & F1 & P & R & F1 & P & R
& F1 & P & R & F1 & P & R & F1 & P & R & F1 & P & R
& (pp) \\
\midrule
Gold
& -- & -- & -- & -- & -- & -- & -- & -- & -- & -- & -- & -- & -- & -- & --
& 86.5 & 80.7 & 96.5
& 51.5 & 61.2 & 48.5
& 56.2 & 89.1 & 46.4
& 63.0 & 76.8 & 61.3
& 0.0 \\
\midrule
Llama-3.1-8B
& \textbf{57.6} & \textbf{57.8} & 57.4 & 46.8 & \textbf{47.9} & 45.8 & \textbf{50.8} & \textbf{50.7} & 50.8 & 51.9 & \textbf{52.0} & 51.8 & \textbf{51.7} & \textbf{52.1} & 51.2
& 55.3 & 52.1 & 66.9 & 46.0 & 52.6 & 44.6 & 51.9 & 78.7 & 43.6 & 50.7 & 61.9 & 50.5
& 12.3 \\
Ministral-3-8B
& \textbf{57.6} & 54.6 & \textbf{61.1} & \textbf{47.1} & 43.2 & \textbf{51.7} & 49.0 & 45.7 & 52.7 & \textbf{52.1} & 48.8 & 55.7 & 51.4 & 47.9 & \textbf{55.4}
& \textbf{56.7} & \textbf{55.4} & 64.4 & 41.4 & 49.2 & 38.8 & 48.2 & 77.0 & 39.4 & 48.1 & 61.0 & 46.2
& 14.9 \\
Phi-3.5-mini
& 53.3 & 49.2 & 58.2 & 46.3 & 42.1 & 51.4 & 49.7 & 45.6 & \textbf{54.8} & 52.0 & 47.3 & \textbf{57.8} & 50.0 & 45.7 & 55.1
& 55.9 & 52.3 & \textbf{67.3} & \textbf{49.1} & \textbf{55.4} & \textbf{47.0} & \textbf{55.2} & \textbf{80.9} & \textbf{46.4} & \textbf{53.2} & \textbf{63.7} & \textbf{52.5}
& \textbf{9.8} \\
Qwen3-8B
& 56.6 & 56.6 & 56.5 & 46.9 & 45.3 & 48.7 & 50.6 & 50.4 & 50.8 & 51.9 & 50.3 & 53.5 & 51.2 & 50.3 & 52.2
& 55.0 & 54.0 & 63.1 & 42.3 & 50.4 & 39.8 & 49.0 & 76.5 & 41.0 & 48.3 & 60.8 & 46.8
& 14.7 \\
\bottomrule
\end{tabular}%
}
\label{tab:task_consistency_level1}
\end{table}

\paragraph{Setup and motivation.}
To characterize cross-level behavior over the event hierarchy, we conduct hierarchical consistency profiling on an \emph{article-sampled} test subset.
We randomly sample 192 test articles and retain the complete Level-1 structure within each article, including subevents and their associated BCE and SCE units (Table~\ref{tab:sampled_test_hv_stats}).
Table~\ref{tab:task_consistency_level1} reports both task performance and hierarchical consistency on this subset, allowing us to compare agreement with gold labels against consistency between higher-level predictions and the model’s own subevent-level predictions for the same article and actor.

\paragraph{Hierarchical consistency as set-overlap F1.}
For each higher-level instance $(u,a)$ with $u\in\{\text{Art.},\text{BCE},\text{SCE}\}$, let $H(u,a)$ denote the model-predicted \emph{directed} value set at level $u$.
We define a reference set $L_{\cup}(u,a)$ as the union of the model’s directed predictions over the supporting subevents of $u$ for the same actor.
Consistency is computed as the set-overlap $F_1$ between $H(u,a)$ and $L_{\cup}(u,a)$:
\begin{equation}
P_{\text{cons}}(u,a)=\frac{|H(u,a)\cap L_{\cup}(u,a)|}{|H(u,a)|},\qquad
R_{\text{cons}}(u,a)=\frac{|H(u,a)\cap L_{\cup}(u,a)|}{|L_{\cup}(u,a)|},
\end{equation}
\begin{equation}
F_{\text{cons}}(u,a)=\frac{2\cdot P_{\text{cons}}(u,a)\cdot R_{\text{cons}}(u,a)}
{P_{\text{cons}}(u,a)+R_{\text{cons}}(u,a)}.
\end{equation}

\paragraph{Results and analysis.}
The \textbf{Gold} results in Table~\ref{tab:task_consistency_level1} show a clear asymmetry across levels.
At the \textbf{Article} level, consistency recall is higher than precision (96.5 vs.\ 80.7), whereas at the \textbf{BCE} and \textbf{SCE} levels, recall is lower than precision (48.5 vs.\ 61.2 for BCE; 46.4 vs.\ 89.1 for SCE).
This pattern is partly shaped by the annotation procedure: after subevent-level human verification, some higher-level labels were further checked with LLM assistance, and article-level annotations were more likely to absorb value information from multiple supporting subevents.
As a result, article-level labels tend to be more inclusive, while BCE and SCE labels remain more selective.

Against this reference, the four models are relatively similar in task performance, with Overall Micro-F1 around 50--52\%, but differ more clearly in hierarchical consistency.
\textbf{Phi-3.5-mini} achieves the highest Overall consistency F1 (53.2\%, $\Delta$Overall = 9.8pp), followed by \textbf{Llama-3.1-8B} (50.7\%, $\Delta$Overall = 12.3pp), while \textbf{Ministral-3-8B} and \textbf{Qwen3-8B} are lower (48.1\% and 48.3\%, with $\Delta$Overall = 14.9pp and 14.7pp).
This indicates that even when models achieve similar unit-level accuracy, they can differ in how well they retain value information when moving from localized subevents to broader contextual units.

\subsection{Controlled Grouping Test (True vs.\ Random, fixed $k$)}
\label{sec:eval3-controlled-grouping}

To test whether the proposed BCE and SCE units provide meaningful composite context rather than arbitrary collections of subevents, we compare gold composite instances against matched random groupings.
The core idea is simple: if the subevents grouped in a gold BCE/SCE form a more useful contextual unit for value understanding, then models should recognize human values more effectively from these \textsc{True} composites than from randomly assembled ones.
Statistics of the evaluated Level-1 instances are summarized in Table~\ref{tab:sampled_test_hv_stats} (\textbf{CGT} rows).

\paragraph{Construction.}
For each gold BCE or SCE instance $(u,a)$, the \textsc{True} condition uses its original composite text.
The \textsc{Random} condition keeps the same actor $a$ and the same group size $k(u)=|\mathcal{P}^{(g)}(u)|$, but replaces the original linked subevent set with a randomly sampled set
$\mathcal{R}^{(g)}_{\text{rand}}(u,a)\subseteq \mathcal{E}^{(g)}(a)$ satisfying
\begin{equation}
|\mathcal{R}^{(g)}_{\text{rand}}(u,a)| = k(u),\quad
\mathcal{R}^{(g)}_{\text{rand}}(u,a) \neq \mathcal{P}^{(g)}(u).
\label{eq:rand-group-simplified}
\end{equation}
The sampled subevent summaries are concatenated to form a pseudo composite input.
Both conditions are evaluated for the same $(u,a)$ pair against the same gold labels.
This comparison tests whether the original BCE/SCE grouping provides a more useful contextual unit for value recognition than random aggregation under matched actor and size constraints.

\paragraph{Evaluation.}
Let $\mathcal{M}$ denote an evaluation metric (e.g., Micro-F1) over all evaluated BCE/SCE instances.
We report
\begin{equation}
\Delta \mathcal{M} = \mathcal{M}_{\textsc{True}} - \mathcal{M}_{\textsc{Random}}.
\label{eq:controlled-gain-simplified}
\end{equation}
A positive $\Delta \mathcal{M}$ indicates that the dataset-defined composite grouping is more helpful for value recognition than a size-matched random grouping.

\begin{figure}[t]
    \centering
    \includegraphics[width=\linewidth]{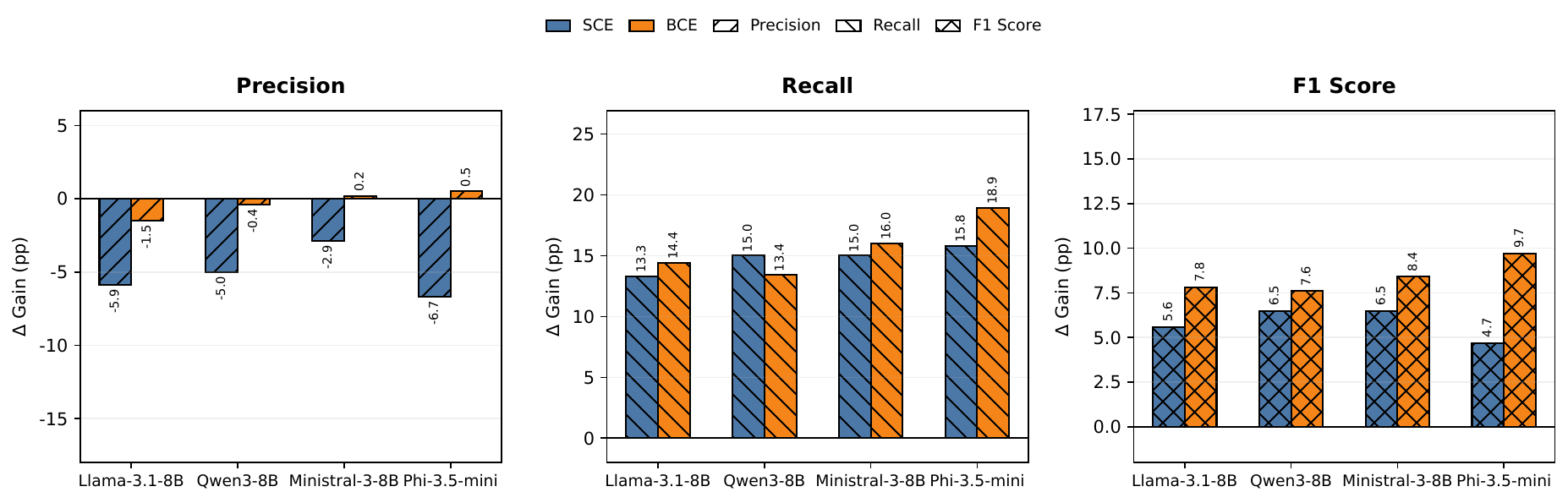}
    \caption{Controlled Grouping Test (Level-1): Precision/Recall/F1 gains of \textsc{True} composites over \textsc{Random} composites (\(\Delta=\textsc{True}-\textsc{Random}\)) for SCE and BCE. The \textsc{Random} condition matches the same actor and the same group size \(k\) by sampling and concatenating \(k\) actor-associated subevents.}
    \label{fig:eval3_delta_prf_l1}
\end{figure}

\paragraph{Results and analysis.}
Fig.~\ref{fig:eval3_delta_prf_l1} reports the controlled grouping gains $\Delta \mathcal{M}$ for Precision/Recall/F1 at the SCE and BCE levels.
Across all models, \textsc{True} composites consistently outperform \textsc{Random} composites, with the gains mainly coming from recall.
This suggests that the dataset-defined BCE/SCE units are meaningful rather than arbitrary: grouping related subevents into composite contextual units makes human value recognition easier than using random pseudo-composites.

Notably, BCE shows stronger and cleaner gains than SCE.
For BCE, recall improvements are substantial (\(+13.4\) to \(+18.9\) pp) while precision is approximately preserved (from \(-1.5\) to \(+0.5\) pp), yielding larger F1 gains (\(+7.6\) to \(+9.7\) pp).
For SCE, recall gains remain strong (\(+13.3\) to \(+15.8\) pp), but precision drops more noticeably (\(-2.9\) to \(-6.7\) pp), producing moderate F1 gains (\(+4.7\) to \(+6.5\) pp).
This suggests that tighter composite grouping may provide more concentrated value-relevant context, whereas broader grouping can improve coverage but also introduce more loosely related associations.
Overall, the uniformly positive \(\Delta\)F1 across models supports that the dataset-provided composite units contribute useful value-relevant context beyond size-matched random grouping.

\subsection{Case Study}
\begin{tcolorbox}[
  enhanced,
  breakable,
  enhanced jigsaw,
  colback=gray!5,
  colframe=gray!40,
  title={Case Study}
]
\textbf{Case 0 (BCE): Father responsibility violation}

\textbf{Task.} Actor-conditioned human value prediction for \textbf{Father} at \textbf{BCE} level.

\vspace{2pt}
\textbf{Unit (BCE).}
Teresa is forced to become the family’s peacemaker and a second mother because both parents are emotionally unavailable.
She blames her father for not intervening in her mother’s alcoholism, and their relationship deteriorates.

\vspace{2pt}
\textbf{Evidence (quotes).}
(i) ``My dad wasn’t approachable, and my mom wasn’t emotionally available.''
(ii) ``I became the peacemaker ... But it was an impossible burden.''
(iii) ``My relationship with my father also suffered.''

\vspace{2pt}
\textbf{Context (surrounding sentences).}
(i) ``In the chaos of addiction, children often take on roles to survive.''
(ii) ``I felt like I had to solve everything on my own.''
(iii) ``I blamed him for allowing my mom to continue her behavior and for not doing anything for us.''

\vspace{2pt}
\textbf{Gold.} \textbf{Contradictory: Be responsible}

\vspace{2pt}
\textbf{Predictions.}
\begin{itemize}[leftmargin=*, itemsep=2pt, label=\textbullet]\setlength\itemsep{1pt}
  \item  \textbf{G1 (Gemini 2.5 Pro):}
  Contradictory: Be responsible, Have the own family secured, Be loving, Be helpful

  \item \textbf{G2 (Llama-3.1-8B):}
  Aligned: Be helpful, Be honest, Be forgiving, Be loving, Be responsible, Have loyalty towards friends
  Contradictory: Be humble

  \item \textbf{G3 (LoRA-tuned Llama 3.1-8B):}
  Contradictory: Be responsible
\end{itemize}

\vspace{2pt}
\tcbline

\textbf{Case 1 (Subevent): Hagia Sophia prayer scene}

\vspace{2pt}
\textbf{Task.} Actor-conditioned human value prediction for \textbf{Religious Authorities and Muslim Worshippers} at \textbf{subevent} level.

\vspace{2pt}
\textbf{Unit (Subevent).}
Worshippers arrived holding prayer rugs as verses from the Quran were read over loudspeakers.

\vspace{2pt}
\textbf{Evidence (quote).}
(i) ``However, it is a place where people from all religions can come and visit as a cultural heritage of all humanity.''

\vspace{2pt}
\textbf{Context (surrounding sentences).}
(i) ``Now this place [the Hagia Sophia] has returned to its original form, it became a mosque again,''
(ii) ``Hagia Sophia served as a church for 916 years ... and a mosque from 1453 to 1934 ...''

\vspace{2pt}
\textbf{Gold.} \textbf{Aligned: Be holding religious faith}

\vspace{2pt}
\textbf{Predictions.}
\begin{itemize}[leftmargin=*, itemsep=2pt, label=\textbullet]\setlength\itemsep{1pt}
  \item \textbf{G1 (Gemini 2.5 Pro):} Aligned: Have a sense of belonging, Be respecting traditions, Be holding religious faith
  \item \textbf{G2 (Llama 3.1-8B):} Aligned: Be holding religious faith, Be helpful, Be honest
  \item \textbf{G3 (LoRA-tuned Llama 3.1-8B):} Aligned: Be holding religious faith.
\end{itemize}
\end{tcolorbox}

\vspace{4pt}
\noindent\textbf{Analysis (Cross-model error patterns).}
The two cases expose two recurring failure modes when predicting \emph{actor-conditioned} human values: 
(\textit{i}) \emph{role-prior bias} that ignores negative context and flips the value direction, and 
(\textit{ii}) \emph{value granularity confusion} that over-expands into adjacent or generic values.

\paragraph{Case 0 (BCE: Father responsibility violation).}
Although the actor token \emph{Father} is semantically associated with positive role expectations (e.g., protection, care, responsibility),
the local narrative depicts an \emph{explicit responsibility breach}: the father is emotionally unavailable and fails to intervene in the mother’s alcoholism, 
forcing the child to assume an ``impossible burden'' and damaging the relationship.
This requires the model to ground the prediction in \emph{contextualized behaviors} rather than a dictionary-style role meaning.

G1 correctly recognizes the \emph{contradictory} direction for \emph{Be responsible}, but shows \emph{semantic inflation} by adding several adjacent family-related values 
(\emph{Have the own family secured}, \emph{Be loving}, \emph{Be helpful}), suggesting insufficient granularity control and weak selectivity under a negative context.
In contrast, G2 exhibits a stronger \emph{role-prior} effect: it places \emph{Be responsible} and other prosocial values under the \emph{aligned} direction, 
which indicates that the model defaults to a ``good parent'' template and underweights the evidence of neglect and non-intervention.
After LoRA fine-tuning, G3 corrects both issues: it recovers the correct \emph{contradictory} label and reduces irrelevant spillover, 
implying improved context sensitivity and better calibration of value selectivity at the BCE level.

\paragraph{Case 1 (Subevent: Hagia Sophia prayer scene).}
This case illustrates \emph{ValueGranularityConfusion}: the gold label is the specific value \emph{Be holding religious faith}, 
but models may drift toward nearby, broader values.
G1 includes \emph{Have a sense of belonging} and \emph{Be respecting traditions} alongside the gold value, reflecting a tendency to conflate faith practice with 
community identity and cultural tradition.
G2 similarly over-generalizes by adding generic moral traits (\emph{Be helpful}, \emph{Be honest}) that are not directly observable in the subevent.
G3 retains only the gold value, suggesting that fine-tuning helps the model stay at the intended semantic granularity and avoid adjacent-value over-association.

\subsection{Genre-wise Robustness}

\paragraph{News-type (genre) categorization.}
We categorize each collected news article into an IPTC-style genre taxonomy following the IPTC NewsCodes genre definitions\footnote{\url{https://cv.iptc.org/newscodes/genre/}}.
Specifically, each article is assigned to one of the following genres:
\textit{Current}, \textit{Feature}, \textit{Background\_Analysis}, \textit{Opinion},
\textit{Special\_Report}, \textit{Profile\_Biography}, \textit{Review}, and \textit{Attribution\_QA}.
The dataset is dominated by three genres (\textit{Current}, \textit{Background\_Analysis}, and \textit{Feature}),
which account for 86.1\% of articles and 84.9\% of value instances.
We therefore restrict the genre-wise analysis to these three most frequent genres.
To ensure fair comparison across genres, we construct a genre-balanced test set by matching unit counts at each hierarchical level:
284 articles, 844 subevents, 428 BCEs, and 410 SCEs per genre.

\begin{figure}[t]
    \centering
    \includegraphics[width=0.98\linewidth]{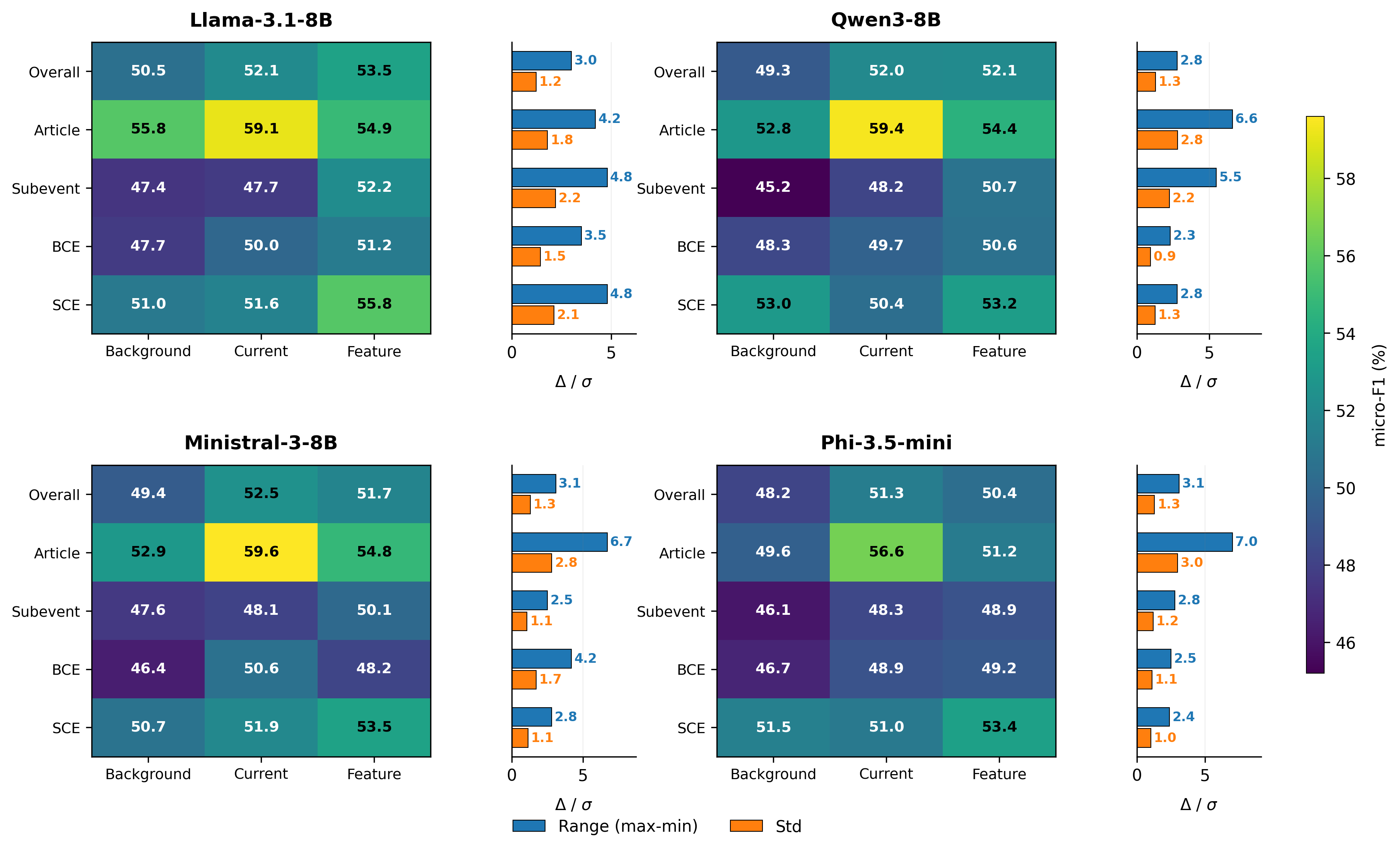}
    \caption{Eval4 news-type comparison (Level-1): Heatmaps show micro-F1 (\%) across hierarchical units and news types, while the right bars quantify genre sensitivity using Range (max--min) and Std over the three genres.}
    \label{fig:eval4_newstype_heatmap_robustness}
\end{figure}

\label{sec:eval4-newstype}

\paragraph{Analysis.}
Figure~\ref{fig:eval4_newstype_heatmap_robustness} shows that \textbf{Current} news consistently yields the strongest performance across all four models.
The largest gains appear at the \textbf{Article} level, reaching up to $\sim$7 points over \textbf{Background\_Analysis}, which suggests that genre differences affect document-level value recognition most strongly.

The \textbf{Article} level is also the most genre-sensitive unit overall, with consistently larger cross-genre variation than the other units.
This likely reflects differences in how the three genres present value-relevant information.
\textbf{Current} news usually reports concrete events more directly, \textbf{Background\_Analysis} tends to embed value cues in explanation and context, and \textbf{Feature} articles often include richer narrative detail that can make full-article judgment less focused.

Once the input is reorganized into \textbf{Subevent}, \textbf{BCE}, or \textbf{SCE} units, genre effects become smaller because predictions rely less on full-document writing style and more on localized or grouped event context.
This helps explain why finer-grained and composite units are relatively more stable across genres.

\section{Limitations}
\label{sec:limitations}

NEVU has several limitations.
First, due to limited annotation budget, our directed human value labels are produced via an LLM-assisted pipeline with targeted manual auditing, and thus may contain residual noise, especially when value cues are implicit, distributed, or direction-sensitive (Table~\ref{tab:dataset_quality_full}).
Second, value inference in news depends critically on actor abstraction and attribution; some instances still involve ambiguous actor definitions or weak grounding, which can propagate uncertainty to value judgments, particularly for short, locally scoped units.
Third, the label distribution is long-tailed: as shown in Fig.~\ref{fig:hv_l1_freq_54_levels}, a subset of Level-1 values remains relatively sparse across units, which may limit robust evaluation and learning for rare value types.

Beyond dataset construction, the current scope of NEVU is also limited.
While NEVU supports fine-grained benchmarking of news-domain value recognition, our present analyses remain focused on the benchmark setting itself.
We do not yet provide a comprehensive study of how value understanding benefits downstream news applications, such as value-aware generation, retrieval, framing analysis, or decision support, nor do we quantify its utility beyond the benchmark tasks.
Finally, news is increasingly multimodal (images, videos, and graphics), whereas NEVU is currently text-only and does not model multimodal cues that may convey normative signals or affect attribution and interpretation.
We view these limitations as important directions for future work, including expanding manual adjudication, strengthening actor canonicalization and grounding, improving coverage of long-tail values, developing downstream task suites, and extending NEVU to multimodal news settings.

\section{Conclusion}
\label{sec:conclusion}

We introduced \textbf{NEVU}, a benchmark dataset for evaluating \emph{actor-conditioned} and \emph{direction-aware} human value recognition in factual news.
NEVU fills an important gap in prior value resources by grounding value prediction in a hierarchy of multi-level event units (\textit{Article}, \textit{Subevent}, \textit{BCE}, and \textit{SCE}) and by annotating directed values over \mbox{(unit, actor)} pairs.
The resulting benchmark provides a fine-grained testbed for assessing whether LLMs can identify human values in news, distinguish aligned from contradictory expressions, and operate consistently across multiple semantic levels.

Across experiments, we find that news-domain human value recognition remains challenging for both proprietary and open-source LLMs under a unified prompting setup.
At the same time, NEVU exhibits clear trainability: lightweight supervised adaptation substantially improves open models on fine-grained value and direction recognition, in some cases approaching or surpassing proprietary baselines.
Our additional analyses further show that multi-level event structure is useful not only for localized evaluation, but also for analyzing how value signals are preserved, grouped, and affected across semantic scopes, as reflected in cross-level consistency, controlled grouping, and robustness under genre variation.
Manual quality assessment further indicates that the benchmark provides reliable actor abstraction and evidence-grounded event structuring, while value correctness remains the most difficult dimension under strict criteria.

Overall, NEVU establishes a new benchmark setting for fine-grained human value evaluation in factual news and provides a foundation for future work on stronger actor grounding, more reliable direction inference, and broader value-aware analysis of news-domain texts.

\section*{Acknowledgements}
\label{sec:acknowledgements}
This work was supported by JST BOOST (Grant Number: JPMJBS2414) and partially supported by JSPS Grant-in-Aid for Scientific Research (Grant Number: 25K03231) and ROIS NII Open Collaborative Research 2025 (Grant Number: 252S4-23623).

\appendix

\section{Data Availability}
\label{app:data_code_availability}

A partial public release of the NEVU dataset is available on HuggingFace:
\url{https://huggingface.co/datasets/WangYao-GoGoGo/NEVU}. To preserve the integrity of the NTCIR-19 FEHU shared-task evaluation, the reserved evaluation portion is excluded from the current public release. The full dataset will be released after the completion of the NTCIR-19 shared task in September 2026.

\section{Human Value Taxonomy}
\label{app:value_hierarchy}
\begin{figure}[t]
    \centering
    \includegraphics[width=\linewidth]{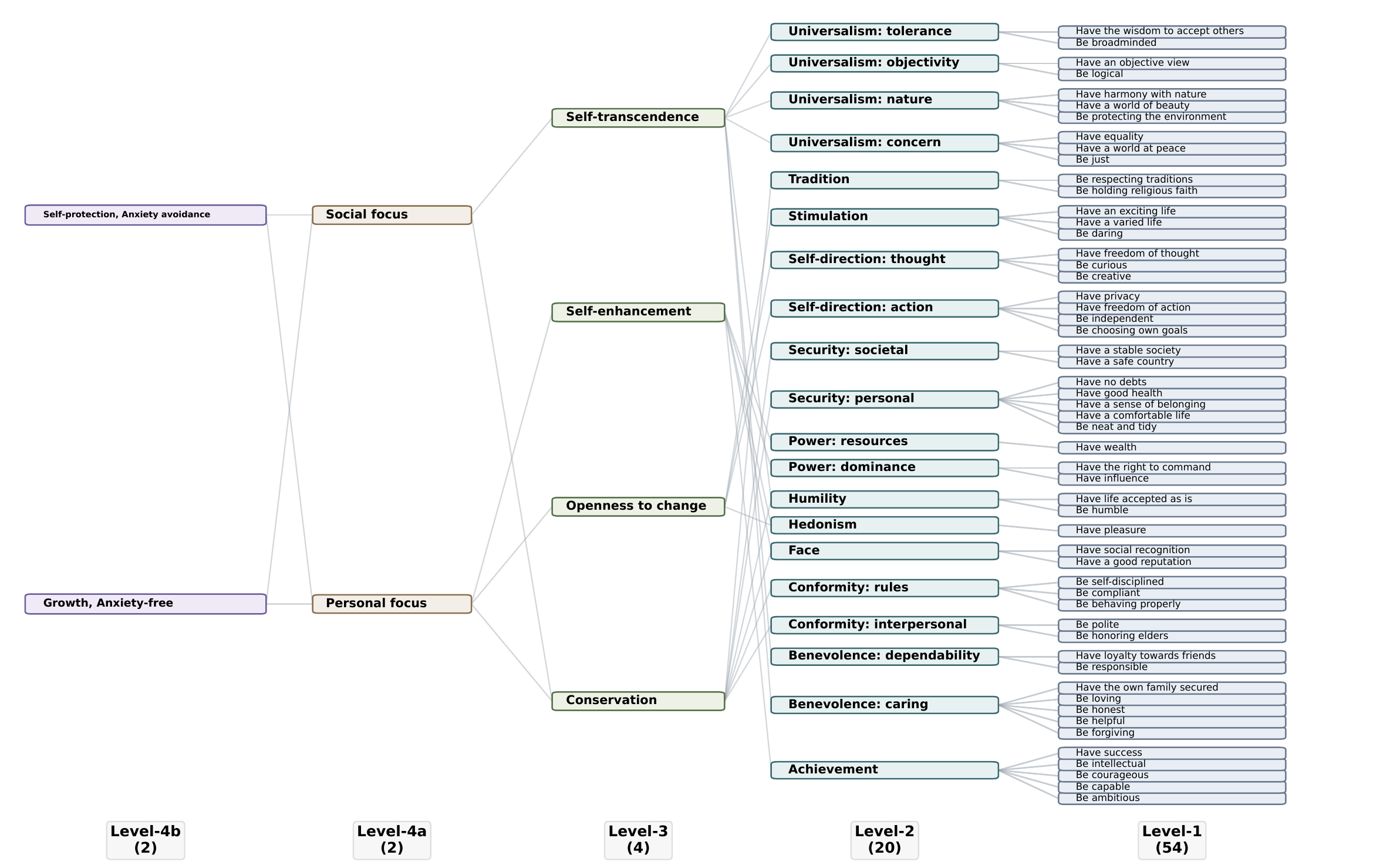}
    \caption{Full hierarchy of the human value taxonomy adopted in this work,
reproduced from \cite{kiesel2022identifying}.
The taxonomy consists of 54 Level-1 values grouped into 20 Level-2 categories,
aligned with four Level-3 motivational domains, and characterized by two
orthogonal Level-4 dimensions: focus (Level-4a: personal / social) and
anxiety regulation (Level-4b: growth/anxiety-free / self-protection/anxiety-avoidance).}
    \label{fig:value_hierarchy_tree}
\end{figure}
Figure~\ref{fig:value_hierarchy_tree} shows the full hierarchy of the adopted
human value taxonomy from \cite{kiesel2022identifying}.
It includes the 54 fine-grained Level-1 values, their grouping into 20 Level-2
categories, the four Level-3 motivational domains, and the two higher-level
dimensions: Level-4a (personal / social focus) and Level-4b (growth /
self-protection with respect to anxiety regulation).

\begin{table}[t]
\centering
\small
\setlength{\tabcolsep}{4.5pt}
\renewcommand{\arraystretch}{1.02}
\caption{External source domains referenced by Wikinews pages in our collection (canonicalized domains).}
\label{tab:app_source_domains}
\begin{tabular}{lllll}
\toprule
\multicolumn{5}{c}{Source domains} \\
\midrule
bbc.com & theguardian.com & nytimes.com & eff.org & greenmatters.com \\
lifehack.org & grist.org & psychologytoday.com & frugalwoods.com & bbc.co.uk \\
lovepanky.com & pickthebrain.com & securityweek.com & krebsonsecurity.com & markmanson.net \\
affordanything.com & tinybuddha.com & getrichslowly.org & kasperskycontenthub.com & brenebrown.com \\
sustainability-times.com & earth911.com & mindful.org & jamesclear.com & konmari.com \\
positivityblog.com & cupofjo.com & goodfinancialcents.com & schneier.com & thesimpledollar.com \\
blog.konmari.com & hbr.org &  &  &  \\
\bottomrule
\end{tabular}
\end{table}


\section{Detailed Data Collection Procedure and Source Composition}
\label{app:phase1_data_collection_sources}

To construct a news corpus that covers \emph{diverse news types}, maintains \emph{temporal diversity}, and yields a \emph{more balanced distribution of human values}, we adopted a three-stage collection strategy:
(i) building a diverse base corpus from an existing dataset,
(ii) value-driven supplementation and crawling to extend coverage, and
(iii) unified filtering and cleaning to ensure consistent validity and safety.
Figure~\ref{fig:phase1_flow} summarizes this Phase-1 pipeline.

\paragraph{Primary sources.}
We collected English news articles from two primary sources:
(1) the publicly available Uknow dataset~\cite{gong2024uknow}, and
(2) English Wikinews stories (2022--2025) accessed via the MediaWiki Action API\footnote{\url{https://en.wikinews.org/w/api.php}.}.
For Wikinews items, we additionally record the canonicalized domains of external links cited on each Wikinews page as provenance signals.
Table~\ref{tab:app_source_domains} lists the observed external source domains (and domains used for supplementary crawling when applicable).

\begin{figure}[t]
    \centering
    \includegraphics[width=\linewidth]{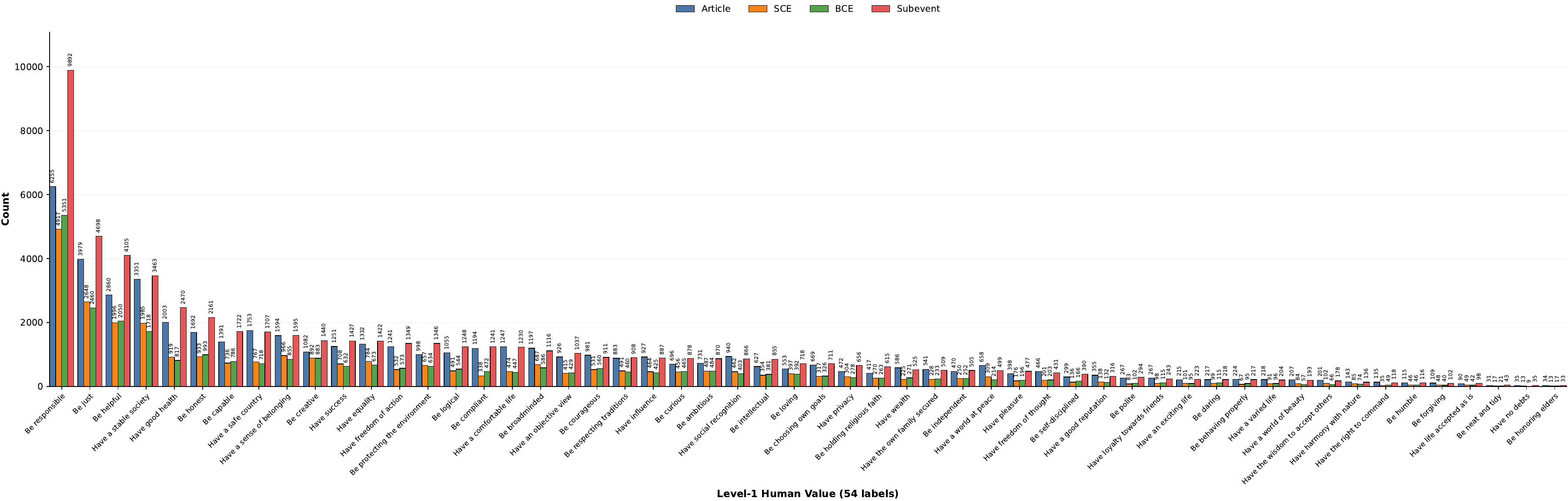}
    \caption{Frequency distribution of the 54 Level-1 human values across Article, Subevent, BCE, and SCE. For each value label, bars compare its count at the four semantic levels; labels are ordered by overall frequency.}
    \label{fig:hv_l1_freq_54_levels}
\end{figure}

\paragraph{Stage 1: Base corpus construction from Uknow.}
We first built a base corpus from Uknow to ensure broad coverage of news types and topics.
Specifically, we used a Transformer-based classifier provided by Knowledgator\footnote{\url{https://docs.knowledgator.com/docs}.} to assign topic and news-type labels to each article, and then performed coarse stratified sampling by news type and publication year, yielding a candidate set of articles.

\paragraph{Stage 2: Value-driven supplementation.}
After obtaining the base candidate set, we conducted preliminary LLM-based value identification to estimate the distribution of human value labels.
We then identified underrepresented value types (e.g., values with fewer than 100 instances) and used OpenAI models to generate semantically related retrieval keywords for targeted supplementation (e.g., for \emph{Be protecting the environment}, keywords include \emph{environment pollution} and \emph{land desertification}).
In parallel, we curated diverse news-source domains (Table~\ref{tab:app_source_domains}) to broaden genre coverage, particularly \emph{soft news} (human-interest/lifestyle), which is often helpful for collecting long-tail value expressions.
Guided by the value-driven keywords and curated domains, we further crawled recent English Wikinews stories from 2022--2025 via the MediaWiki Action API to enrich temporal diversity and mitigate year-wise concentration, yielding 1,110 additional articles.

\paragraph{Stage 3: Unified filtering and cleaning.}
Regardless of whether candidate articles originated from Uknow (Stage~1) or the Wikinews crawl (Stage~2), we applied a \emph{unified filtering and cleaning pipeline}, implemented via a \emph{single-pass GPT-4-nano evaluation}, to ensure consistent standards of validity, narrative suitability, and safety across all sources.
A single prompt jointly assessed multiple criteria for each article, and the resulting signals were used for filtering:
(1) \textbf{Category validity and sensitivity assessment} (excluding highly sensitive narratives where value interpretation may vary substantially across perspectives);
(2) \textbf{Human-centered narrative suitability} (retaining articles that foreground human actions, social interactions, or moral framing);
(3) \textbf{Safety and discrimination screening}; and
(4) \textbf{Cleaning and normalization} (deduplication, removal of corrupted/overly short texts, and encoding/format repair).

In total, our corpus contains 2,865 articles (1,755 from Uknow and 1,110 from Wikinews 2022--2025), which serves as the foundation for our event-centric human value understanding dataset.

\section{Detailed Schema and Construction Rules for the Event Knowledge Base}
\label{app:event_kb_details}

For reproducibility, we provide a detailed description of the event knowledge base schema and construction rules used in Phase-2.
To make the formal definitions easier to interpret, we explain them together with the running example shown in Fig.~\ref{fig:phase2_flow}.

For each article, we assign a GUID $g\in\mathcal{G}$ and denote the full article text, together with metadata such as title, time, and source, as $x^{(g)}$.
We further segment $x^{(g)}$ into an ordered sentence sequence
$\mathbf{S}^{(g)}=\big(s^{(g)}_{i}\big)_{i=1}^{N_g}$,
where $i$ is a \texttt{sentence\_id}.
In Fig.~\ref{fig:phase2_flow}, the left panel shows the source article $x^{(g)}$ for the Paris police headquarters attack case.
This article--sentence indexing scheme serves as the base layer of traceability for all subsequent units.

Starting from the article-level document $x^{(g)}$, we extract a set of fine-grained subevents,
$\mathcal{E}^{(g)}=\{e^{(g)}_p\}_{p=1}^{n_g}$,
where $p$ is a local \texttt{subevent\_id}.
Each subevent stores a short natural-language summary and is explicitly grounded to a set of supporting \texttt{sentence\_ids} in the source article.
As illustrated in the central panel of Fig.~\ref{fig:phase2_flow}, the Paris attack report is decomposed into local event units such as the attacker entering the building, attacking colleagues, stabbing victims, people being killed or critically injured, police shooting the attacker, authorities sealing off the area, officials visiting the scene, prosecutors launching an investigation, and police searching the attacker's home.
To support consistent characterization of subevents, we use the WikiEvents Ontology~\cite{li2021document} as a conceptual reference and perform open-type event extraction with trigger-like cues as anchors for identifying coherent subevent boundaries and descriptions.
These subevents function as the primary evidence-grounded local units in the benchmark, making later value inference more localizable.

Based on these grounded subevents, we further construct a set of behavior-based composite events,
$\mathcal{B}^{(g)}=\{b^{(g)}_q\}_{q=1}^{B_g}$,
where $q$ is a local \texttt{behavior\_event\_id}.
Each BCE groups a set of closely related subevents into a relatively more complete behavior or response chain.
Operationally, a BCE stores an ordered list of behavior items, and each item may reference the originating \texttt{subevent\_id} for traceability.
On the right side of Fig.~\ref{fig:phase2_flow}, BCE-0 groups the attack sequence, including initiating the attack, stabbing colleagues, and causing deaths and injuries, whereas BCE-1 groups the response sequence, including shooting the attacker, sealing off the area, launching the investigation, and searching the attacker's home.
Compared with isolated subevents, BCEs provide a broader contextual unit when value implications become clearer from linked actions and immediate outcomes viewed together.

At a broader contextual scope, we also construct a set of story-based composite events,
$\mathcal{N}^{(g)}=\{c^{(g)}_r\}_{r=1}^{M_g}$,
where $r$ is a local \texttt{story\_event\_id}.
Each SCE groups a broader set of related subevents into a relatively more complete story episode, optionally accompanied by a short title and narrative description.
Operationally, each SCE stores a set of related \texttt{subevent\_ids} under the same GUID, providing explicit links from the story-level unit back to subevents and, transitively, to sentence evidence.
In the lower-left panel of Fig.~\ref{fig:phase2_flow}, SCE-0 summarizes the attack and its immediate consequences and links to subevents 1--6, while SCE-1 summarizes the law-enforcement and governmental response and links to subevents 7--10.
Compared with isolated subevents, SCEs provide a broader contextual unit for interpreting value-relevant actions, responses, and consequences within the same story segment.

In parallel, we identify a set of major social actors,
$\mathcal{A}^{(g)}=\{a^{(g)}_k\}_{k=1}^{K_g}$,
where $k$ is a local actor index under the same GUID.
To reduce ambiguity from aliases, coreference, and varying mention specificity, we canonicalize actor mentions and, when appropriate, abstract them into role-based actors defined by their event participation, such as initiators and recipients/targets.
For traceability, we additionally store alignments from actor mentions to sentence-level mentions in $\mathbf{S}^{(g)}$.
As shown in the upper-right panel of Fig.~\ref{fig:phase2_flow}, the running example normalizes surface mentions into stable actor entities such as the attacker, law enforcement and authorities, and government and political leaders.
This actor layer is critical for later human value annotation, because the benchmark evaluates actor-conditioned value understanding over grounded $(u,a)$ pairs rather than assigning values to event units in isolation.

Overall, a central design principle of the event knowledge base is explicit cross-level linking across units derived from the same article.
Subevents are directly grounded to \texttt{sentence\_ids}; BCEs reference the \texttt{subevent\_ids} underlying their grouped subevents; SCEs store related \texttt{subevent\_ids} as the basis of each story segment; and actor mentions are aligned back to sentence-level expressions.
This traceability design supports two benchmark goals.
First, it makes value inference more interpretable by allowing higher-level predictions to be traced back to concrete sentence-grounded evidence.
Second, it supports analysis across semantic granularities by enabling comparisons of article-, subevent-, BCE-, and SCE-level value judgments under shared actor and evidence structures.

\section{Detailed Procedure for Human Value Annotation}
\label{app:phase3_details}

Building on the event knowledge base constructed in Phase-2, we perform actor-conditioned human value annotation over a unified set of semantic units.
For each article $g\in\mathcal{G}$, we define
\begin{equation}
\label{eq:unit-set}
\mathcal{U}^{(g)}=\{u^{(g)}_{\mathrm{art}}\}\cup \mathcal{E}^{(g)}\cup \mathcal{B}^{(g)}\cup \mathcal{N}^{(g)}.
\end{equation}
where $u^{(g)}_{\mathrm{art}}$ denotes the article-level unit corresponding to the full text $x^{(g)}$,
$\mathcal{E}^{(g)}$ is the subevent set,
$\mathcal{B}^{(g)}$ is the behavior-based composite event (BCE) set,
and $\mathcal{N}^{(g)}$ is the story-based composite event (SCE) set.
Human values are then annotated over \mbox{(unit, actor)} pairs $(u,a)$, where $u\in\mathcal{U}^{(g)}$ and $a\in\mathcal{A}^{(g)}$.

Figures~\ref{fig:phase2_flow} and~\ref{fig:phase3_hv_examples} together illustrate this process with a running example.
Figure~\ref{fig:phase2_flow} shows how a source article is transformed into linked semantic units and normalized actors, while Fig.~\ref{fig:phase3_hv_examples} shows how value judgments are subsequently assigned to selected \mbox{(unit, actor)} pairs across the Article, Subevent, BCE, and SCE levels.
For example, the same actor may be evaluated in different semantic contexts: a local subevent may express one value cue, whereas a BCE or SCE may reveal a broader pattern of responsibility, justice, or social stability.
This is why the verification procedure operates on grounded \mbox{(unit, actor, value, direction)} candidates rather than on article-level labels alone.

\subsection{Stage 1: LLM-based Human Value Identification with Voting}
\label{app:stage1_llm_voting}

To reduce systematic bias from any single model, we first perform candidate value identification using a heterogeneous set of LLMs
$\mathcal{M}_{\mathrm{id}}=\{\text{GPT-4o},\ \text{Gemini-2.5-Pro},\ \text{DeepSeek-Reasoner}\}$
\cite{openai_gpt4o_system_card_2024,gemini25_2025,deepseek_reasoner_api_2026}.
At this stage, each model independently proposes candidate directed value labels for each $(u,a)$ pair.

In the running example, these candidates may be generated for units at different semantic levels, such as a subevent describing the attacker stabbing colleagues, a BCE representing the broader attack sequence, an SCE summarizing the institutional response, or the full article-level unit.
As illustrated in Fig.~\ref{fig:phase3_hv_examples}, the same actor can be associated with different value judgments depending on the semantic unit under consideration.
For instance, the attacker may be linked to contradictory values such as \emph{Have a safe country}, \emph{Have a stable society}, or \emph{Have good health}, while institutional actors may be linked to aligned values such as \emph{Be responsible}, \emph{Be courageous}, or \emph{Be just}.

We aggregate the model outputs using 3-way voting.
A directed value label is retained as a high-confidence candidate only when all three models agree on the same $(v,d)$ prediction.
All remaining cases are treated as low-confidence labels and forwarded to the subsequent QA-based verification stage.

\subsection{Stage 2: QA-based Verification of Low-confidence Labels}

For low-confidence candidates from Stage 1, we perform a second verification step using a separate QA-oriented model pool
$\mathcal{M}_{\mathrm{qa}}=\{\text{GPT-5},\ \text{Gemini-2.5-Flash},\ \text{Claude-Sonnet-4}\}$
\cite{openai_gpt5_system_card_2025,gemini25_2025,anthropic_sonnet45_system_card_2025}.
Rather than re-predicting labels from scratch, this stage verifies each candidate $(u,a,v,d)$ through a QA-style procedure inspired by CLAVE~\cite{yao2024clave}, with an emphasis on evidence grounding and robustness.

For each candidate, we construct a prompt conditioned on the source article $x^{(g)}$, the target unit $u$ together with its supporting evidence context, the target actor $a$, and the definition of value $v$.
Each model then gives a binary decision (\texttt{Yes}/\texttt{No}) on whether direction $d$ of value $v$ is supported by the evidence for actor $a$ in unit $u$.

The example in Figs.~\ref{fig:phase2_flow} and~\ref{fig:phase3_hv_examples} also illustrates why this second stage is necessary.
Stage 1 is designed to identify candidate human value instances in a relatively open-ended manner, so as to maximize coverage of plausible value cues for each \mbox{(unit, actor)} pair.
By contrast, Stage 2 no longer asks the model to generate labels from scratch; instead, it performs fine-grained QA-style verification for each identified candidate $(u,a,v,d)$.

More specifically, for each candidate value instance, we provide the model with richer verification context, including the source article $x^{(g)}$, the target unit $u$ together with its supporting evidence context, the target actor $a$, and the definition of value $v$.
This design allows the model to make a more precise judgment about whether the proposed value label and its direction are genuinely supported by the event evidence, rather than merely being plausible at a coarse semantic level.

We aggregate the verification results by 3-way voting:
candidates with three \texttt{Yes} votes are accepted,
candidates with three \texttt{No} votes are rejected,
and mixed cases are sent to Stage 3 for human verification.

\subsection{Stage 3: Human Verification}

For candidates that remain unresolved after Stage 2, we conduct targeted human verification.
Annotators use a dedicated interface (Fig.~\ref{fig:bce_annotation_ui}) that presents the target instance together with structured cross-level context, including related Article/Subevent/BCE/SCE units and traceability signals such as sentence-level evidence and composite-event membership links.
The verification procedure follows the staged design shown in Fig.~\ref{fig:phase2_flow}, while Fig.~\ref{fig:phase3_hv_examples} illustrates example gold human-value labels from the running event example, showing how values are represented across different semantic levels.

Within this interface, annotators inspect the target instance in relation to its surrounding semantic structure rather than as an isolated label decision.
They check (i) whether the actor attribution is correct, (ii) whether the assigned value label is appropriate, and (iii) whether the aligned/contradictory direction is supported by the described behaviors and outcomes.
Among the low-confidence instances passed to human verification, only those receiving unanimous acceptance from annotators are retained in the final benchmark annotations.

\section{Additional Dataset Statistics and Label Distribution}
\label{app:dataset_statistics}

Figure~\ref{fig:hv_l1_freq_54_levels} visualizes the frequency distribution of the 54 Level-1 human values across the four semantic granularities.
Several values occur markedly more often than others, indicating that news discourse tends to foreground particular normative themes.
For example, \textit{Be responsible} is consistently among the most frequent labels, suggesting that news narratives often frame actions, policies, and decisions in terms of responsibility and accountability.
Values such as \textit{Be just} and \textit{Be helpful} are also highly frequent, which is broadly consistent with the journalistic emphasis on fairness, social obligation, and harm--benefit consequences.
At the same time, less frequent values remain represented across all levels, supporting evaluation beyond only the most salient and commonly expressed value dimensions.

\section{Human Annotation Interface}
\begin{figure}[t]
  \centering
  \begin{minipage}[t]{0.32\linewidth}
    \centering
    \includegraphics[width=\linewidth]{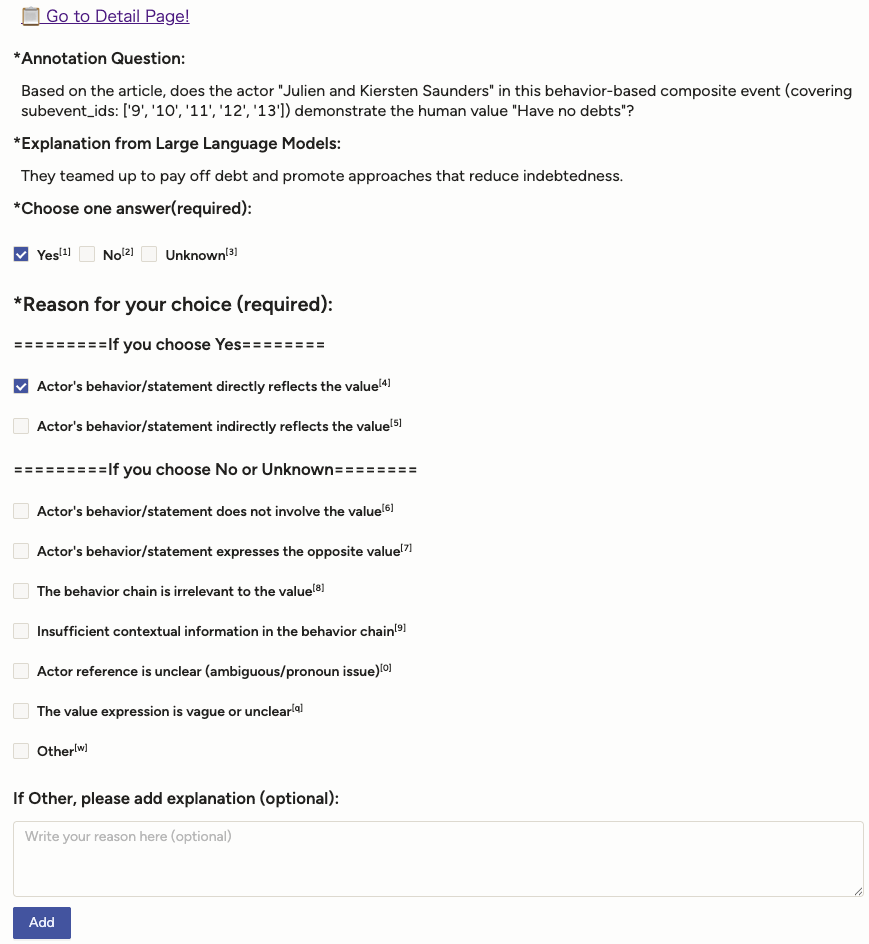}
    \small (a) Annotation page.
  \end{minipage}\hfill
  \begin{minipage}[t]{0.32\linewidth}
    \centering
    \includegraphics[width=\linewidth]{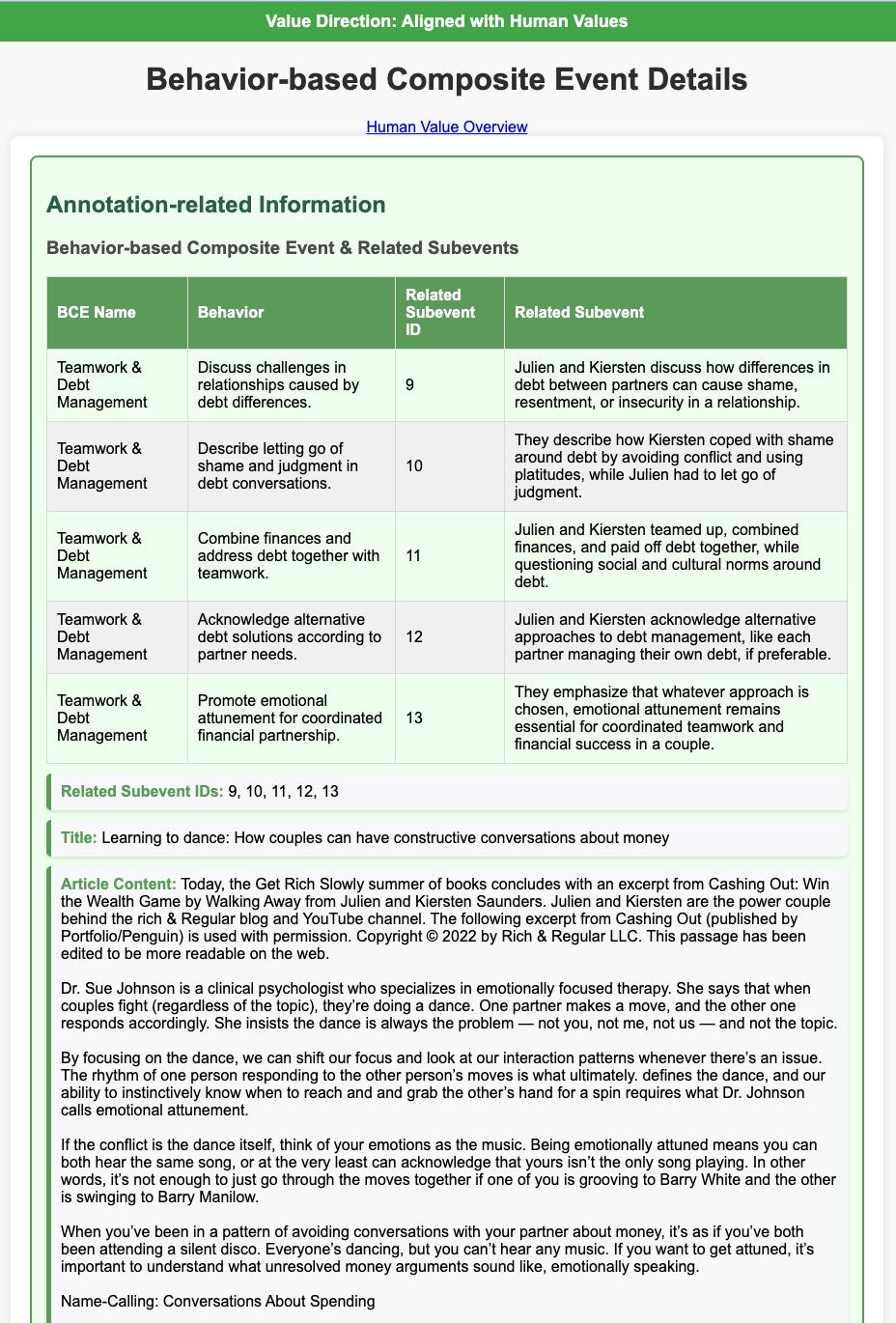}
    \small (b) Data detail page.
  \end{minipage}\hfill
  \begin{minipage}[t]{0.32\linewidth}
    \centering
    \includegraphics[width=\linewidth]{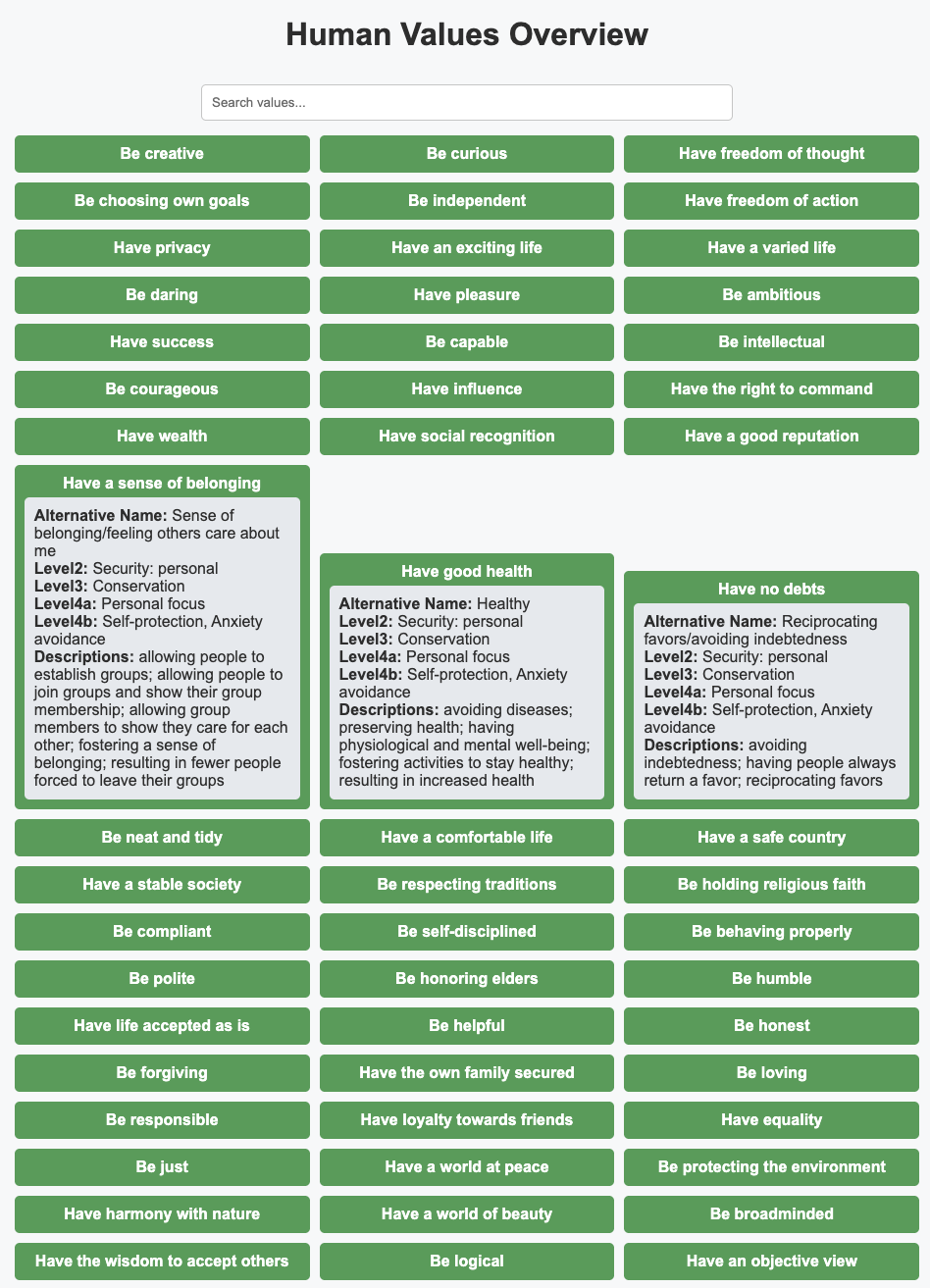}
    \small (c) Human-value overview.
  \end{minipage}

  \caption{Screenshots of our manual annotation interface for a BCE human-value labeling example.}
  \label{fig:bce_annotation_ui}
\end{figure}

This section presents our web-based manual annotation interface with a BCE human-value labeling example (Fig.~\ref{fig:bce_annotation_ui}).
The interface consists of three components:
(a) a human-value labeling page implemented in Label Studio\footnote{\url{https://labelstud.io/}} for actor-conditioned Yes/No/Unknown judgments with selected reasons;
(b) a custom-developed data-detail view that links each BCE to its supporting subevents and article-level evidence for traceable review; and
(c) a custom-developed searchable human-value reference page that provides value definitions and hierarchy information to support consistent label interpretation across annotators.

\section{Manual Quality Evaluation}
\label{app:human_evaluation}

\begin{table*}[t]
\centering
\small
\setlength{\tabcolsep}{5.5pt}
\renewcommand{\arraystretch}{1.08}
\caption{Manual quality evaluation results (Soft/Hard, \%) with N-weighted overall scores.}
\label{tab:dataset_quality_full}
\begin{tabular}{lccccc}
\toprule
\textbf{Dimension} & \textbf{Subevent} & \textbf{BCE} & \textbf{SCE} & \textbf{Article} & \textbf{Overall} \\
\midrule
Subevent Evidence Support   & 91.5/91.0 & --        & --        & --        & -- \\
Aggregation Justification   & --        & 95.7/91.4 & 95.7/91.4 & --        & -- \\
Evidence Inheritance        & --        & 92.1/84.3 & 96.4/92.9 & --        & -- \\
Narrative Coherence         & --        & --        & 95.7/91.4 & --        & -- \\
Local Adjacency Coherence   & --        & 93.6/87.1 & --        & --        & -- \\
Actor Concept Validity      & 92.5/90.0 & 97.1/97.1 & 90.7/88.6 & 98.9/97.8 & \textbf{94.8/93.4} \\
Actor Grounding Adequacy    & 85.5/82.0 & 99.3/98.6 & 89.3/87.1 & 94.4/93.3 & \textbf{92.1/90.3} \\
Human Value Correctness     & 82.0/81.0 & 88.6/88.6 & 82.1/78.6 & 85.6/84.4 & \textbf{84.6/83.2} \\
\midrule
\textbf{Sample size $N$}    & 100       & 70        & 70        & 45        & 285 \\
\bottomrule
\end{tabular}
\end{table*}

We evaluate NEVU from multiple dimensions covering both the reliability of event-centric units and the faithfulness of actor-conditioned human value annotations. 
The evaluation is conducted on a stratified sample across four granularities: \textit{Article} ($N{=}45$), \textit{Subevent} ($N{=}100$), \textit{BCE} ($N{=}70$), and \textit{SCE} ($N{=}70$). 
For each instance, annotators inspect the target unit, the associated abstract social actor, and an evidence package grounded in the source article (e.g., linked sentences and supporting spans).

\paragraph{Dimensions.}
We assess the sample from the following dimensions:
\begin{itemize}[leftmargin=*, itemsep=2pt, label=\textbullet]
  \item \textbf{Actor Concept Validity.} Whether the abstract actor is coherent and identifiable in the article context (allowing synonymous mentions), and is suitable as a stable target for attribution.
  \item \textbf{Actor Grounding Adequacy.} Whether the provided unit contains sufficient actor-attributable evidence to support attribution (participation/being affected/taking action/expressing stance), rather than relying on name appearance.
  \item \textbf{Human Value Correctness.} Whether the evidence in the unit supports the assigned value \emph{direction} (aligned / contradictory) for the actor--unit pair.
  \item \textbf{Subevent Evidence Support.} (\textit{Subevent only}) Whether the subevent’s core fact is explicitly supported by the source article beyond background cues.
  \item \textbf{Composite Evidence Support.} (\textit{BCE/SCE only}) Whether the composite unit is sufficiently supported by its linked subevents and their source grounding.
  \item \textbf{Aggregation Justification.} (\textit{BCE/SCE only}) Whether the grouped subevents form a reasonable composite context without mixing clearly unrelated content or introducing unsupported relations.
  \item \textbf{Local Chain Coherence.} (\textit{BCE only}) Whether the grouped subevents form a coherent local behavior or response chain under the provided order.
  \item \textbf{Narrative Coherence.} (\textit{SCE only}) Whether the grouped subevents form a coherent story segment with interpretable progression across related developments.
\end{itemize}

\paragraph{Annotators and rating scheme.}
We provide detailed definitions and auditing instructions for all quality dimensions to five annotators (one PhD student and four Master’s students). 
Each dimension is rated using a three-way scheme with dimension-specific labels (e.g., \textit{Yes / Cannot judge / No}, \textit{Strong / Mixed / Weak}, \textit{Consistent / Partial / Inconsistent}, \textit{High / Medium / Low}). 
For reporting, the three-way ratings are mapped to percentage scores to enable aggregation and cross-level comparison.

\paragraph{Results.}
Detailed results are shown in Table~\ref{tab:dataset_quality_full}. 
The $N$-weighted overall scores indicate strong dataset reliability, with Actor Concept Validity at 94.8/93.4 (Soft/Hard), Actor Grounding Adequacy at 92.1/90.3, and Human Value Correctness at 84.6/83.2. 
At the level-wise breakdown, the largest drop appears in actor grounding for subevents (85.5/82.0), consistent with shorter units providing fewer referential and attribution cues. 
For composite units, BCE and SCE maintain high aggregation-related scores, suggesting that their grouped event context is generally well supported by linked subevents and rarely introduces unsupported content. 
Human Value Correctness is highest for BCE (88.6/88.6), exceeding subevents (82.0/81.0) and SCEs (82.1/78.6). One plausible explanation is that BCEs often provide a more concentrated composite context over closely related actions and outcomes, whereas SCEs tend to involve broader story segments whose value implications may be more diffuse and context-dependent. 
Overall, these findings support the reliability of NEVU’s actor abstractions and composite event structuring, while also confirming that strict value attribution remains the most challenging aspect of the dataset.

\section{Level-2 Human Value Recognition Results}
\label{app:hvr-l2-discussion}

\begin{table}[t]
\centering
\small
\setlength{\tabcolsep}{2.2pt}
\renewcommand{\arraystretch}{1.08}
\resizebox{\linewidth}{!}{
\begin{tabular}{llccccc ccccc ccccc}
\toprule
\multirow{2}{*}{\textbf{Group}} & \multirow{2}{*}{\textbf{Model}}
& \multicolumn{5}{c}{\textbf{Micro-F1}} 
& \multicolumn{5}{c}{\textbf{Macro-F1}} 
& \multicolumn{5}{c}{\textbf{DRR$\downarrow$}} \\
\cmidrule(lr){3-7}\cmidrule(lr){8-12}\cmidrule(lr){13-17}
& 
& \textbf{Art.} & \textbf{Sub.} & \textbf{BCE} & \textbf{SCE} & \textbf{Overall}
& \textbf{Art.} & \textbf{Sub.} & \textbf{BCE} & \textbf{SCE} & \textbf{Overall}
& \textbf{Art.} & \textbf{Sub.} & \textbf{BCE} & \textbf{SCE} & \textbf{Overall} \\
\midrule
\multirow{9}{*}{G1}
& Claude Haiku 4.5
& 51.18 & 50.30 & 46.76 & 48.62 & 49.42
& 38.18 & 43.01 & 36.36 & 40.82 & 39.71
& 4.39 & 2.10 & 3.28 & 3.09 & 3.46 \\
& Claude Sonnet 4.5
& 53.87 & 53.48 & 50.85 & 51.49 & 52.56
& 40.71 & 47.22 & 40.28 & 42.20 & 42.73
& 3.24 & 1.63 & 2.98 & 3.05 & 2.84 \\
& Claude Opus 4.5
& 55.77 & \underline{55.53} & \underline{53.20} & \underline{54.52} & 54.86
& 45.31 & 50.13 & 43.34 & 44.85 & 46.35
& 2.88 & \underline{1.47} & 2.68 & 2.63 & 2.52 \\
& Deepseek-chat
& 44.27 & 48.46 & 47.74 & 50.22 & 47.15
& 31.57 & 39.41 & 35.83 & 37.90 & 35.61
& 3.15 & 1.98 & 3.06 & 3.16 & 2.91 \\
& Deepseek-reasoner
& 50.45 & 50.81 & 51.21 & 52.41 & 51.12
& 40.25 & 43.55 & 39.84 & 44.37 & 42.26
& 2.08 & 1.94 & 2.19 & 1.83 & 2.02 \\
& OpenAI gpt-4.1
& 51.90 & 51.42 & 49.60 & 51.37 & 51.18
& 39.18 & 43.45 & 39.60 & 42.12 & 40.79
& 3.42 & 1.59 & 2.75 & 2.59 & 2.76 \\
& OpenAI gpt-5.2
& 50.11 & 51.99 & 51.50 & 52.53 & 51.31
& 39.23 & 44.32 & 41.19 & 41.62 & 41.33
& 2.73 & 1.59 & 2.38 & 2.51 & 2.39 \\
& OpenAI o3
& 45.22 & 50.00 & 48.79 & 50.16 & 47.99
& 35.76 & 42.45 & 38.44 & 41.16 & 39.25
& \underline{1.43} & 1.51 & \underline{1.96} & \underline{1.71} & \underline{1.61} \\
& Gemini 2.5 Pro
& \underline{61.26} & 54.07 & 51.71 & 51.49 & \underline{55.41}
& \underline{51.11} & \underline{50.16} & \underline{43.72} & \underline{45.46} & \underline{48.55}
& 4.64 & 2.18 & 3.36 & 2.90 & 3.55 \\
\midrule
\multirow{4}{*}{G2}
& Ministral-3-8B
& \underline{30.56} & \underline{34.78} & \underline{33.64} & \underline{35.11} & \underline{33.12}
& \underline{19.72} & \underline{27.96} & \underline{24.67} & \underline{24.43} & \underline{23.83}
& \underline{0.06} & 0.05 & \underline{0.06} & 0.05 & \underline{0.05} \\
& Phi-3.5-mini
& 22.52 & 15.09 & 17.50 & 18.86 & 19.19
& 12.63 & 9.24 & 9.94 & 12.18 & 11.34
& \underline{0.06} & \underline{0.04} & \underline{0.04} & \underline{0.04} & \underline{0.05} \\
& Qwen3-8B
& 27.84 & 23.61 & 24.91 & 25.22 & 25.79
& 14.75 & 14.57 & 14.53 & 16.51 & 15.27
& 5.19 & 1.47 & 2.57 & 2.48 & 3.39 \\
& Llama-3.1-8B
& 25.10 & 19.43 & 20.15 & 20.60 & 22.11
& 11.97 & 10.38 & 9.38 & 11.48 & 11.34
& 12.57 & 5.11 & 5.77 & 6.02 & 8.43 \\
\midrule
\multirow{4}{*}{G3}
& Llama-3.1-8B
& \underline{\textbf{64.66}} & \underline{\textbf{60.74}} & 59.20 & \underline{\textbf{60.84}} & \underline{\textbf{62.13}}
& 52.32 & \underline{\textbf{53.23}} & \underline{\textbf{53.33}} & 51.24 & \underline{\textbf{53.10}}
& 4.20 & 0.83 & 1.13 & 1.37 & 2.35 \\
& Qwen3-8B
& 63.72 & 59.04 & \underline{\textbf{60.23}} & 60.80 & 61.61
& 52.04 & 51.36 & 50.96 & \underline{\textbf{52.11}} & 52.45
& \underline{\textbf{0.05}} & \underline{\textbf{0.01}} & \underline{\textbf{0.01}} & \underline{\textbf{0.01}} & \underline{\textbf{0.03}} \\
& Ministral-3-8B
& 63.89 & 57.83 & 56.82 & 59.58 & 60.55
& \underline{\textbf{52.87}} & 51.12 & 45.82 & 50.45 & 51.33
& \underline{\textbf{0.05}} & \underline{\textbf{0.01}} & 0.02 & \underline{\textbf{0.01}} & \underline{\textbf{0.03}} \\
& Phi-3.5-mini
& 61.58 & 56.21 & 58.28 & 60.38 & 59.73
& 48.64 & 50.40 & 47.59 & 49.31 & 50.13
& 0.07 & \underline{\textbf{0.01}} & 0.02 & 0.02 & 0.04 \\
\bottomrule
\end{tabular}
}
\caption{Human Value Recognition results on 20 coarse-grained Level-2 human values.
G1: proprietary API LLMs; G2: open-weight LLMs (prompting only); G3: open-weight LLMs with LoRA fine-tuning.
Best-in-group is underlined; overall best is underlined in bold.}
\label{app:hvr-l2-merged-levels}
\end{table}

Table~\ref{app:hvr-l2-merged-levels} reports the \textbf{Level-2} results (20 coarse-grained values) across semantic levels (Article/Subevent/BCE/SCE) and Overall.
Level-2 labels are deterministically mapped from Level-1 outputs, therefore the table mainly serves as a \emph{coarse-grained} view of the same predictions.

\paragraph{Overall comparison.}
At Level-2, \textbf{Gemini 2.5 Pro} remains the strongest proprietary baseline in \textbf{G1} with \textbf{55.41} Overall Micro-F1 and \textbf{48.55} Overall Macro-F1.
The best results are achieved by \textbf{G3} LoRA-adapted models, where \textbf{Llama-3.1-8B+LoRA} reaches \textbf{62.13}/\textbf{53.10} (Micro/Macro) Overall, followed closely by \textbf{Qwen3-8B+LoRA} (\textbf{61.61}/\textbf{52.45}).

\paragraph{Granularity robustness and direction errors.}
Across groups, Level-2 performance is generally more stable across BCE/SCE than Level-1, reflecting fewer near-miss confusions after coarse mapping.
For direction errors, several \textbf{G3} models achieve very low DRR on composite units (often around \textbf{0.01} on Sub/BCE/SCE), while \textbf{Llama-3.1-8B+LoRA} keeps a higher Overall DRR (\textbf{2.35}) despite the best F1, indicating that direction reversal remains a non-trivial error mode even under the coarse taxonomy.

\section{Dataset Construction Cost}
\label{app:human_verification_cost}

\paragraph{Annotator recruitment.}
To support the human verification stage, we recruited 32 student annotators, including 13 undergraduate, 12 master's, and 7 doctoral students, spanning five national backgrounds. The verification campaign lasted approximately 2.5 months.

\paragraph{Human effort and cost.}
The human verification stage involved a multi-annotator auditing campaign over approximately 2.5 months, with contributions from both paid annotators and volunteers.
In total, this targeted verification effort amounted to 455 person-hours.
Because human review was reserved for cases not confidently resolved by the earlier automatic stages, this figure reflects focused auditing effort rather than exhaustive manual annotation of all released value instances.
The paid human verification cost was approximately \$2.5k USD.

\paragraph{API cost.}
LLM API usage for event construction, label proposal, and QA-style verification cost approximately \$3.2k USD, covering OpenAI, Gemini, DeepSeek, and Claude.

\paragraph{Overall cost.}
The combined cost of paid human verification and LLM API usage was approximately \$5.6k USD.


\begin{thebibliography}{8}

\bibitem{schwartz2012refining}
Schwartz, S., Cieciuch, J., Vecchione, M., Davidov, E., Fischer, R., Beierlein, C., Ramos, A., Verkasalo, M., Lönnqvist, J., Demirutku, K., et al. Refining the theory of basic individual values. {\em Journal Of Personality And Social Psychology}. \textbf{103}, 663 (2012)

\bibitem{rokeach1973nature}
Rokeach, M. The nature of human values. (Free press,1973)

\bibitem{brown2002life}
Brown, D. \& Crace, R. Life values inventory: Facilitator’s guide. {\em Williamsburg, VA}. (2002)

\bibitem{haerpfer2020world}
Haerpfer, C., Inglehart, R., Moreno, A., Welzel, C., Kizilova, K., Diez-Medrano, J., Lagos, M., Norris, P., Ponarin, E., Puranen, B., et al.
\newblock World Values Survey: Round Seven---Country-Pooled Datafile.
\newblock {\em Madrid, Spain \& Vienna, Austria: JD Systems Institute \& WVSA Secretariat}, 7 pp., 2021 (2020).

\bibitem{entman1993framing}
Entman, R. M.
\newblock Framing: Toward clarification of a fractured paradigm.
\newblock {\em Journal of Communication} \textbf{43}(4), 51--58 (1993).

\bibitem{devreese2005news}
de Vreese, C. H.
News framing: Theory and typology.
\emph{Information Design Journal + Document Design}, \textbf{13}(1), 51--62 (2005).

\bibitem{semetko2000framing}
Semetko, H. A. \& Valkenburg, P. M.
Framing European politics: A content analysis of press and television news.
\emph{Journal of Communication}, \textbf{50}(2), 93--109 (2000).

\bibitem{hu2024bad}Hu, B., Sheng, Q., Cao, J., Shi, Y., Li, Y., Wang, D. \& Qi, P. Bad actor, good advisor: Exploring the role of large language models in fake news detection. {\em Proceedings Of The AAAI Conference On Artificial Intelligence}. \textbf{38}, 22105-22113 (2024)
\bibitem{pratelli2025evaluation}Pratelli, M., Bianchi, J., Pinelli, F. \& Petrocchi, M. Evaluation of Reliability Criteria for News Publishers with Large Language Models. {\em Proceedings Of The 17th ACM Web Science Conference 2025}. pp. 179-188 (2025)

\bibitem{zhang2025systematic}Zhang, H., Yu, P. \& Zhang, J. A systematic survey of text summarization: From statistical methods to large language models. {\em ACM Computing Surveys}. \textbf{57}, 1-41 (2025)

\bibitem{yang2025accuracy}Yang, K. \& Menczer, F. Accuracy and political bias of news source credibility ratings by large language models. {\em Proceedings Of The 17th ACM Web Science Conference 2025}. pp. 127-137 (2025)

\bibitem{cao2025survey}Cao, Y., Li, S., Liu, Y., Yan, Z., Dai, Y., Yu, P. \& Sun, L. A survey of ai-generated content (aigc). {\em ACM Computing Surveys}. \textbf{57}, 1-38 (2025)

\bibitem{yao2023instructions}Yao, J., Yi, X., Wang, X., Wang, J. \& Xie, X. From Instructions to Intrinsic Human Values–A Survey of Alignment Goals for Big Models. {\em ArXiv Preprint ArXiv:2308.12014}. (2023)

\bibitem{rezapour2025tales}Rezapour, R., Jeoung, S., You, Z. \& Diesner, J. Tales of Morality: Comparing Human-and LLM-Generated Moral Stories from Visual Cues. {\em Findings Of The Association For Computational Linguistics: EMNLP 2025}. pp. 18915-18933 (2025)
\bibitem{yang2025constructing}Yang, Z., Jian, P., Li, C., Wang, C., Zhang, X. \& Lu, W. Constructing Your Model’s Value Distinction: Towards LLM Alignment with Anchor Words Tuning. {\em Findings Of The Association For Computational Linguistics: EMNLP 2025}. pp. 5932-5948 (2025)

\bibitem{su2025understanding}Su, Y., Zhang, J., Yang, S., Wang, X., Hu, L. \& Wang, D. Understanding how value neurons shape the generation of specified values in llms. {\em ArXiv Preprint ArXiv:2505.17712}. (2025)

\bibitem{bai2022constitutional}Bai, Y., Kadavath, S., Kundu, S., Askell, A., Kernion, J., Jones, A., Chen, A., Goldie, A., Mirhoseini, A., McKinnon, C., et al. Constitutional ai: Harmlessness from ai feedback. {\em ArXiv Preprint ArXiv:2212.08073}. (2022)
\bibitem{ji2023beavertails}Ji, J., Liu, M., Dai, J., Pan, X., Zhang, C., Bian, C., Chen, B., Sun, R., Wang, Y. \& Yang, Y. Beavertails: Towards improved safety alignment of llm via a human-preference dataset. {\em Advances In Neural Information Processing Systems}. \textbf{36} pp. 24678-24704 (2023)
\bibitem{guo2025counterfactual}Guo, H., Yao, J., Zhou, X., Yi, X. \& Xie, X. Counterfactual Reasoning for Steerable Pluralistic Value Alignment of Large Language Models. {\em ArXiv Preprint ArXiv:2510.18526}. (2025)

\bibitem{kiesel2022identifying}
Kiesel, J., Alshomary, M., Handke, N., Cai, X., Wachsmuth, H. \& Stein, B. Identifying the human values behind arguments. {\em Proceedings Of The 60th Annual Meeting Of The Association For Computational Linguistics (Volume 1: Long Papers)}. pp. 4459-4471 (2022)

\bibitem{mirzakhmedova2024touche23}Mirzakhmedova, N., Kiesel, J., Alshomary, M., Heinrich, M., Handke, N., Cai, X., Barriere, V., Dastgheib, D., Ghahroodi, O., SadraeiJavaheri, M., et al. The touché23-valueeval dataset for identifying human values behind arguments. {\em Proceedings Of The 2024 Joint International Conference On Computational Linguistics, Language Resources And Evaluation (LREC-COLING 2024)}. pp. 16121-16134 (2024)

\bibitem{chen2025mova}Chen, Z., Sun, J., Li, C., Nguyen, T., Yao, J., Yi, X., Xie, X., Tan, C. \& Xie, L. MoVa: Towards generalizable classification of human morals and values. {\em Proceedings Of The 2025 Conference On Empirical Methods In Natural Language Processing}. pp. 33204-33248 (2025)

\bibitem{marcuzzo2025morables}Marcuzzo, M., Zangari, A., Albarelli, A., Camacho-Collados, J. \& Pilehvar, M. MORABLES: A Benchmark for Assessing Abstract Moral Reasoning in LLMs with Fables. {\em Proceedings Of The 2025 Conference On Empirical Methods In Natural Language Processing}. pp. 27715-27739 (2025)

\bibitem{lourie2021scruplescorpuscommunityethical}Lourie, N., Bras, R. \& Choi, Y. Scruples: A Corpus of Community Ethical Judgments on 32,000 Real-Life Anecdotes.  (2021), https://arxiv.org/abs/2008.09094

\bibitem{zhang2024moka}Zhang, X., Wu, W., Beauchamp, N. \& Wang, L. MOKA: Moral knowledge augmentation for moral event extraction. {\em Proceedings Of The 2024 Conference Of The North American Chapter Of The Association For Computational Linguistics: Human Language Technologies (Volume 1: Long Papers)}. pp. 4481-4502 (2024)

\bibitem{becker2025moralization}Becker, M., Sommer, M., Tapken, L., Teh, Y. \& Brocai, B. The Moralization Corpus: Frame-Based Annotation and Analysis of Moralizing Speech Acts across Diverse Text Genres. {\em ArXiv Preprint ArXiv:2512.15248}. (2025)

\bibitem{shen2025valuecompass}Shen, H., Knearem, T., Ghosh, R., Yang, Y., Clark, N., Mitra, T. \& Huang, Y. Valuecompass: A framework for measuring contextual value alignment between human and llms. {\em Proceedings Of The 9th Widening NLP Workshop}. pp. 75-86 (2025)

\bibitem{qiu2022valuenet}
Qiu, L., Zhao, Y., Li, J., Lu, P., Peng, B., Gao, J. \& Zhu, S. Valuenet: A new dataset for human value driven dialogue system. {\em Proceedings Of The AAAI Conference On Artificial Intelligence}. \textbf{36}, 11183-11191 (2022)

\bibitem{yao2024clave}Yao, J., Yi, X. \& Xie, X. Clave: An adaptive framework for evaluating values of llm generated responses. {\em Advances In Neural Information Processing Systems}. \textbf{37} pp. 58868-58900 (2024)

\bibitem{trager2025mftcxplain}Trager, J., Vargas, F., Alves, D., Guida, M., Ngueajio, M., Agrawal, A., Daryani, Y., Karimi-Malekabadi, F. \& Arco, F. MFTCXplain: A Multilingual Benchmark Dataset for Evaluating the Moral Reasoning of LLMs through Multi-hop Hate Speech Explanation. {\em Findings Of The Association For Computational Linguistics: EMNLP 2025}. pp. 15709-15740 (2025)

\bibitem{bosselut2019comet}Bosselut, A., Rashkin, H., Sap, M., Malaviya, C., Celikyilmaz, A. \& Choi, Y. COMET: Commonsense transformers for automatic knowledge graph construction. {\em ACL}. (2019)
\bibitem{hwang2021comet}Hwang, J., Bhagavatula, C., Le Bras, R., Da, J., Sakaguchi, K., Bosselut, A. \& Choi, Y. COMET-ATOMIC 2020: On symbolic and neural commonsense knowledge graphs. {\em AAAI}. (2021)

\bibitem{kiesel2023semeval}
Kiesel, J., Alshomary, M., Mirzakhmedova, N., Heinrich, M., Handke, N., Wachsmuth, H. \& Stein, B. Semeval-2023 task 4: Valueeval: Identification of human values behind arguments. {\em Proceedings Of The 17th International Workshop On Semantic Evaluation (SemEval-2023)}. pp. 2287-2303 (2023)

\bibitem{sorensen2024value}Sorensen, T., Jiang, L., Hwang, J., Levine, S., Pyatkin, V., West, P., Dziri, N., Lu, X., Rao, K., Bhagavatula, C., et al. Value kaleidoscope: Engaging ai with pluralistic human values, rights, and duties. {\em Proceedings Of The AAAI Conference On Artificial Intelligence}. \textbf{38}, 19937-19947 (2024)

\bibitem{yao2024value}
Yao, J., Yi, X., Gong, Y., Wang, X. \& Xie, X. Value FULCRA: Mapping Large Language Models to the Multidimensional Spectrum of Basic Human Value. {\em Proceedings Of The 2024 Conference Of The North American Chapter Of The Association For Computational Linguistics: Human Language Technologies (Volume 1: Long Papers)}. pp. 8754-8777 (2024)

\bibitem{ren2024valuebench}
Ren, Y., Ye, H., Fang, H., Zhang, X. \& Song, G. ValueBench: Towards Comprehensively Evaluating Value Orientations and Understanding of Large Language Models. {\em Proceedings Of The 62nd Annual Meeting Of The Association For Computational Linguistics (Volume 1: Long Papers)}. pp. 2015-2040 (2024)

\bibitem{norhashim2024measuring}
Norhashim, H. \& Hahn, J. Measuring Human-AI Value Alignment in Large Language Models. {\em Proceedings Of The AAAI/ACM Conference On AI, Ethics, And Society}. \textbf{7} pp. 1063-1073 (2024)

\bibitem{gong2024uknow}Gong, B., Tan, S., Feng, Y., Xie, X., Li, Y., Chen, C., Zheng, K., Shen, Y. \& Zhao, D. UKnow: A Unified Knowledge Protocol with Multimodal Knowledge Graph Datasets for Reasoning and Vision-Language Pre-Training. {\em Advances In Neural Information Processing Systems}. \textbf{37} pp. 9612-9633 (2024)

\bibitem{li2021document}Li, S., Ji, H. \& Han, J. Document-Level Event Argument Extraction by Conditional Generation. {\em Proceedings Of The 2021 Conference Of The North American Chapter Of The Association For Computational Linguistics: Human Language Technologies}. pp. 894-908 (2021)

\bibitem{chakravarthy1994composite}Chakravarthy, S., Krishnaprasad, V., Anwar, E. \& Kim, S. Composite events for active databases: Semantics, contexts and detection. {\em VLDB}. \textbf{94} pp. 606-617 (1994)

\bibitem{keith2023survey}Keith Norambuena, B., Mitra, T. \& North, C. A survey on event-based news narrative extraction. {\em ACM Computing Surveys}. \textbf{55}, 1-39 (2023)

\bibitem{li2021future}Li, M., Li, S., Wang, Z., Huang, L., Cho, K., Ji, H., Han, J. \& Voss, C. The Future is not One-dimensional: Complex Event Schema Induction by Graph Modeling for Event Prediction. {\em Proceedings Of The 2021 Conference On Empirical Methods In Natural Language Processing}. pp. 5203-5215 (2021)

\bibitem{zhang2024analyzing}Zhang, Z., Cao, Y., Ye, C., Ma, Y., Liao, L. \& Chua, T. Analyzing Temporal Complex Events with Large Language Models? A Benchmark towards Temporal, Long Context Understanding. {\em Proceedings Of The 62nd Annual Meeting Of The Association For Computational Linguistics (Volume 1: Long Papers)}. pp. 1588-1606 (2024)

\bibitem{zhou2025llm}
Zhou, Z., Li, C., Chen, X., Wang, S., Chao, Y., Li, Z., Wang, H., Shi, Q., Tan, Z., Han, X., \emph{et al.}
\newblock LLM$\times$MapReduce: Simplified long-sequence processing using large language models.
\newblock In \emph{Proceedings of the 63rd Annual Meeting of the Association for Computational Linguistics
(Volume 1: Long Papers)}, pages 27664--27678, 2025.

\bibitem{10.1016/j.neucom.2025.130170}Wang, Y., Tao, Z., Chang, H., Huang, N., Jin, L. \& Luo, X. Multimodal understanding of human values in videos: A benchmark dataset and PLM-based method. {\em Neurocomput.}. \textbf{638} (2025,7)

\bibitem{gretz2019largescaledatasetargumentquality}Gretz, S., Friedman, R., Cohen-Karlik, E., Toledo, A., Lahav, D., Aharonov, R. \& Slonim, N. A Large-scale Dataset for Argument Quality Ranking: Construction and Analysis.  (2019)

\bibitem{rashkin-etal-2019-towards}Rashkin, H., Smith, E., Li, M. \& Boureau, Y. Towards Empathetic Open-domain Conversation Models: A New Benchmark and Dataset. {\em Proceedings Of The 57th Annual Meeting Of The Association For Computational Linguistics}. pp. 5370-5381 (2019,7)

\bibitem{perez-etal-2022-red}Perez, E., Huang, S., Song, F., Cai, T., Ring, R., Aslanides, J., Glaese, A., McAleese, N. \& Irving, G. Red Teaming Language Models with Language Models. {\em Proceedings Of The 2022 Conference On Empirical Methods In Natural Language Processing}. pp. 3419-3448 (2022,12)

\bibitem{bai2022traininghelpfulharmlessassistant}Bai, Y., Jones, A., Ndousse, K., Askell, A., Chen, A., DasSarma, N., Drain, D., Fort, S., Ganguli, D., Henighan, T., Joseph, N., Kadavath, S., Kernion, J., Conerly, T., El-Showk, S., Elhage, N., Hatfield-Dodds, Z., Hernandez, D., Hume, T., Johnston, S., Kravec, S., Lovitt, L., Nanda, N., Olsson, C., Amodei, D., Brown, T., Clark, J., McCandlish, S., Olah, C., Mann, B. \& Kaplan, J. Training a Helpful and Harmless Assistant with Reinforcement Learning from Human Feedback. (2022)

\bibitem{fraser-etal-2022-moral}Fraser, K., Kiritchenko, S. \& Balkir, E. Does Moral Code have a Moral Code? Probing Delphi's Moral Philosophy. {\em Proceedings Of The 2nd Workshop On Trustworthy Natural Language Processing (TrustNLP 2022)}. pp. 26-42 (2022,7), https://aclanthology.org/2022.trustnlp-1.3/

\bibitem{karra2022estimating}Karra, S., Nguyen, S. \& Tulabandhula, T. Estimating the personality of white-box language models. {\em ArXiv Preprint ArXiv:2204.12000}. (2022)

\bibitem{caron2022identifying}Caron, G. \& Srivastava, S. Identifying and manipulating the personality traits of language models. {\em ArXiv Preprint ArXiv:2212.10276}. (2022)
\bibitem{li2023does}Li, X., Li, Y., Joty, S., Liu, L., Huang, F., Qiu, L. \& Bing, L. Does Gpt-3 demonstrate psychopathy. {\em Evaluating Large Language Models From A Psychological Perspective}. (2023)
\bibitem{miotto2022gpt}Miotto, M., Rossberg, N. \& Kleinberg, B. Who is GPT-3? An exploration of personality, values and demographics. {\em Proceedings Of The Fifth Workshop On Natural Language Processing And Computational Social Science (NLP+ CSS)}. pp. 218-227 (2022)
\bibitem{rao-etal-2023-chatgpt}Rao, H., Leung, C. \& Miao, C. Can ChatGPT Assess Human Personalities? A General Evaluation Framework. {\em Findings Of The Association For Computational Linguistics: EMNLP 2023}. pp. 1184-1194 (2023,12), https://aclanthology.org/2023.findings-emnlp.84/
\bibitem{jiang2023evaluating}Jiang, G., Xu, M., Zhu, S., Han, W., Zhang, C. \& Zhu, Y. Evaluating and inducing personality in pre-trained language models. {\em Advances In Neural Information Processing Systems}. \textbf{36} pp. 10622-10643 (2023)
\bibitem{song2023have}Song, X., Gupta, A., Mohebbizadeh, K., Hu, S. \& Singh, A. Have large language models developed a personality?: Applicability of self-assessment tests in measuring personality in llms. {\em ArXiv Preprint ArXiv:2305.14693}.
\bibitem{zhang2023measuring}Zhang, Z., Bai, F., Gao, J. \& Yang, Y. Measuring value understanding in language models through discriminator-critique gap.  (2023)
\bibitem{zhang2023heterogeneous}Zhang, Z., Liu, N., Qi, S., Zhang, C., Rong, Z., Yang, Y. \& Cui, S. Heterogeneous value evaluation for large language models. {\em ArXiv Preprint ArXiv:2305.17147}. (2023)
\bibitem{pan2023llms}Pan, K. \& Zeng, Y. Do llms possess a personality? making the mbti test an amazing evaluation for large language models. {\em ArXiv Preprint ArXiv:2307.16180}. (2023)
\bibitem{serapio2023personality}Serapio-García, G., Safdari, M., Crepy, C., Sun, L., Fitz, S., Romero, P., Abdulhai, M., Faust, A. \& Matarić, M. Personality Traits in Large Language Models.  (2025), https://arxiv.org/abs/2307.00184
\bibitem{ganesan2023systematic}
Ganesan, A., Lal, Y., Nilsson, A. \& Schwartz, H.
\newblock Systematic evaluation of GPT-3 for zero-shot personality estimation.
\newblock {\em Proceedings of the 13th Workshop on Computational Approaches to Subjectivity, Sentiment, \& Social Media Analysis}, pp.~390--400, 2023.
\bibitem{huang-etal-2024-reliability}Huang, J., Jiao, W., Lam, M., Li, E., Wang, W. \& Lyu, M. On the Reliability of Psychological Scales on Large Language Models. {\em Proceedings Of The 2024 Conference On Empirical Methods In Natural Language Processing}. pp. 6152-6173 (2024,11), https://aclanthology.org/2024.emnlp-main.354/
\bibitem{abdulhai2024moral}Abdulhai, M., Serapio-Garcia, G., Crepy, C., Valter, D., Canny, J. \& Jaques, N. Moral foundations of large language models. {\em Proceedings Of The 2024 Conference On Empirical Methods In Natural Language Processing}. pp. 17737-17752 (2024)
\bibitem{simmons2023moral}Simmons, G. Moral mimicry: Large language models produce moral rationalizations tailored to political identity. {\em Proceedings Of The 61st Annual Meeting Of The Association For Computational Linguistics (Volume 4: Student Research Workshop)}. pp. 282-297 (2023)
\bibitem{scherrer2023evaluating}Scherrer, N., Shi, C., Feder, A. \& Blei, D. Evaluating the moral beliefs encoded in llms. {\em Advances In Neural Information Processing Systems}. \textbf{36} pp. 51778-51809 (2023)
\bibitem{bodroza2023personality}Bodroza, B., Dinic, B. \& Bojic, L. Personality testing of GPT-3: limited temporal reliability, but highlighted social desirability of GPT-3's personality instruments results. {\em ArXiv E-prints}. pp. arXiv-2306 (2023)
\bibitem{la2025open}La Cava, L. \& Tagarelli, A. Open models, closed minds? on agents capabilities in mimicking human personalities through open large language models. {\em Proceedings Of The AAAI Conference On Artificial Intelligence}. \textbf{39}, 1355-1363 (2025)

\bibitem{jiang2021can}Jiang, L., Hwang, J., Bhagavatula, C., Bras, R., Liang, J., Dodge, J., Sakaguchi, K., Forbes, M., Borchardt, J., Gabriel, S., et al. Can machines learn morality? the delphi experiment. {\em ArXiv Preprint ArXiv:2110.07574}. (2021)

\bibitem{ayyubi2024beyond}Ayyubi, H., Thomas, C., Chum, L., Lokesh, R., Chen, L., Niu, Y., Lin, X., Feng, X., Koo, J., Ray, S., et al. Beyond Grounding: Extracting Fine-Grained Event Hierarchies Across Modalities. {\em Proceedings Of The AAAI Conference On Artificial Intelligence}. \textbf{38}, 17664-17672 (2024)

\bibitem{huang2024textee}Huang, K., Hsu, I., Parekh, T., Xie, Z., Zhang, Z., Natarajan, P., Chang, K., Peng, N. \& Ji, H. TextEE: Benchmark, reevaluation, reflections, and future challenges in event extraction. {\em Findings Of The Association For Computational Linguistics: ACL 2024}. pp. 12804-12825 (2024)

\bibitem{chen2020event}Chen, X. \& Li, Q. Event modeling and mining: a long journey toward explainable events: X. Chen, Q. Li. {\em The VLDB Journal}. \textbf{29}, 459-482 (2020)

\bibitem{liu2025utilizing}Liu, W., Zeng, D., Zhou, L., Chen, W., Zhang, M., Liu, D., He, X. \& Li, H. Utilizing Contextual Clues and Role Correlations for Enhancing Document-level Event Argument Extraction. {\em IEEE Transactions On Audio, Speech And Language Processing}. (2025)

\bibitem{chen2023cheer}Chen, M., Cao, Y., Zhang, Y. \& Liu, Z. CHEER: Centrality-aware high-order event reasoning network for document-level event causality identification. {\em Proceedings Of The 61st Annual Meeting Of The Association For Computational Linguistics (Volume 1: Long Papers)}. pp. 10804-10816 (2023)

\bibitem{zheng2024comprehensive}Zheng, H., Wang, S. \& Huang, L. A comprehensive survey on document-level information extraction. {\em Proceedings Of The Workshop On The Future Of Event Detection (FuturED)}. pp. 58-72 (2024)

\bibitem{guan2022event}Guan, S., Cheng, X., Bai, L., Zhang, F., Li, Z., Zeng, Y., Jin, X. \& Guo, J. What is event knowledge graph: A survey. {\em IEEE Transactions On Knowledge And Data Engineering}. \textbf{35}, 7569-7589 (2022)

\bibitem{jin2022event}Jin, X., Li, M. \& Ji, H. Event schema induction with double graph autoencoders. {\em Proceedings Of The 2022 Conference Of The North American Chapter Of The Association For Computational Linguistics: Human Language Technologies}. pp. 2013-2025 (2022)

\bibitem{moore-etal-2024-large}Moore, J., Deshpande, T. \& Yang, D. Are Large Language Models Consistent over Value-laden Questions?. {\em Findings Of The Association For Computational Linguistics: EMNLP 2024}. pp. 15185-15221 (2024,11), https://aclanthology.org/2024.findings-emnlp.891/

\bibitem{ozeki-etal-2025-normative}Ozeki, K., Ando, R., Morishita, T., Abe, H., Mineshima, K. \& Okada, M. Normative Reasoning in Large Language Models: A Comparative Benchmark from Logical and Modal Perspectives. {\em Proceedings Of The 8th BlackboxNLP Workshop: Analyzing And Interpreting Neural Networks For NLP}. pp. 276-294 (2025,11), https://aclanthology.org/2025.blackboxnlp-1.17/

\bibitem{sachdeva2025normative}Sachdeva, P. \& Nuenen, T. Normative evaluation of large language models with everyday moral dilemmas. {\em Proceedings Of The 2025 ACM Conference On Fairness, Accountability, And Transparency}. pp. 690-709 (2025)

\bibitem{wang2023aligning}Wang, Y., Zhong, W., Li, L., Mi, F., Zeng, X., Huang, W., Shang, L., Jiang, X. \& Liu, Q. Aligning large language models with human: A survey. {\em ArXiv Preprint ArXiv:2307.12966}. (2023)

\bibitem{chakraborty2025structured}Chakraborty, M., Wang, L. \& Jurgens, D. Structured moral reasoning in language models: A value-grounded evaluation framework. {\em Proceedings Of The 2025 Conference On Empirical Methods In Natural Language Processing}. pp. 30283-30311 (2025)

\bibitem{zhang2022directquote}Zhang, Y. \& Liu, Y. DirectQuote: A dataset for direct quotation extraction and attribution in news articles. {\em Proceedings Of The Thirteenth Language Resources And Evaluation Conference}. pp. 6959-6966 (2022)

\bibitem{glockner2025neoqa}Glockner, M., Jiang, X., Ribeiro, L., Gurevych, I. \& Dreyer, M. Neoqa: Evidence-based question answering with generated news events. {\em Findings Of The Association For Computational Linguistics: ACL 2025}. pp. 11842-11926 (2025)

\bibitem{zhang2024survey}Zhang, B., Dai, G., Niu, F., Yin, N., Fan, X., Wang, S., Cao, X. \& Huang, H. A survey of stance detection on social media: New directions and perspectives. {\em ArXiv Preprint ArXiv:2409.15690}. (2024)

\bibitem{nguyen2024news}
Nguyen, D. \& Hekman, E.
\newblock The news framing of artificial intelligence: a critical exploration of how media discourses make sense of automation.
\newblock {\em AI \& Society} \textbf{39}, 437--451 (2024).

\bibitem{knutson2024news}Knutson, B., Hsu, T., Ko, M. \& Tsai, J. News source bias and sentiment on social media. {\em PloS One}. \textbf{19}, e0305148 (2024)

\bibitem{kmainasi2025llamalens}Kmainasi, M., Shahroor, A., Hasanain, M., Laskar, S., Hassan, N. \& Alam, F. LlamaLens: Specialized multilingual llm for analyzing news and social media content. {\em Findings Of The Association For Computational Linguistics: NAACL 2025}. pp. 5627-5649 (2025)

\bibitem{askell2021general}Askell, A., Bai, Y., Chen, A., Drain, D., Ganguli, D., Henighan, T., Jones, A., Joseph, N., Mann, B., DasSarma, N., et al. A general language assistant as a laboratory for alignment. {\em ArXiv Preprint ArXiv:2112.00861}. (2021)

\bibitem{openai_gpt52_model_2025}
OpenAI.
\newblock GPT-5.2 Model.
\newblock 2025.
\newblock \url{https://developers.openai.com/api/docs/models/gpt-5.2}.
\newblock Accessed: 2026-03-08.

\bibitem{openai_gpt4_2024}
OpenAI, Josh Achiam, et al.
\newblock GPT-4 Technical Report.
\newblock 2024.
\newblock \url{https://arxiv.org/abs/2303.08774}.

\bibitem{openai_o3_o4mini_release_2025}OpenAI Introducing OpenAI o3 and o4-mini.  (2025), https://openai.com/index/introducing-o3-and-o4-mini/, Accessed: 2026-03-08
\bibitem{anthropic_haiku45_system_card_2025}Anthropic Claude Haiku 4.5 System Card.  (2025), https://www.anthropic.com/claude-haiku-4-5-system-card, Accessed: 2026-03-08
\bibitem{anthropic_sonnet45_system_card_2025}Anthropic Claude Sonnet 4.5 System Card.  (2025), https://www.anthropic.com/claude-sonnet-4-5-system-card, Accessed: 2026-03-08
\bibitem{anthropic_opus45_system_card_2025}Anthropic Claude Opus 4.5 System Card.  (2025), https://www.anthropic.com/claude-opus-4-5-system-card, Accessed: 2026-03-08
\bibitem{gemini25_2025}Comanici, G., Bieber, E., et al. Gemini 2.5: Pushing the Frontier with Advanced Reasoning, Multimodality, Long Context, and Next Generation Agentic Capabilities. {\em ArXiv Preprint ArXiv:2507.06261}. (2025), https://arxiv.org/abs/2507.06261
\bibitem{deepseekv32_2025}
DeepSeek-AI.
\newblock DeepSeek-V3.2: Pushing the Frontier of Open Large Language Models.
\newblock {\em arXiv preprint arXiv:2512.02556}, 2025.
\newblock \url{https://arxiv.org/abs/2512.02556}.
\bibitem{deepseek_reasoner_api_2026}
DeepSeek.
Reasoning Model (deepseek-reasoner).
(2026),
\url{https://api-docs.deepseek.com/guides/reasoning_model},
Accessed: 2026-03-08.
\bibitem{weerawardhena2025llama31foundationaisecurityllm8binstructtechnicalreport}Weerawardhena, S., Kassianik, P., et al. Llama-3.1-FoundationAI-SecurityLLM-8B-Instruct Technical Report.  (2025), https://arxiv.org/abs/2508.01059
\bibitem{yang2025qwen3technicalreport}Yang, A., Li, A., et al. Qwen3 Technical Report.  (2025), https://arxiv.org/abs/2505.09388
\bibitem{liu2026ministral3}Liu, A., Khandelwal, K., et al. Ministral 3.  (2026), https://arxiv.org/abs/2601.08584
\bibitem{abdin2024phi3technicalreporthighly}Abdin, M., Aneja, J., et al. Phi-3 Technical Report: A Highly Capable Language Model Locally on Your Phone.  (2024), https://arxiv.org/abs/2404.14219
\bibitem{openai_gpt4o_system_card_2024}OpenAI GPT-4o System Card.  (2024), https://openai.com/index/gpt-4o-system-card/, Accessed: 2026-03-08
\bibitem{openai_gpt5_system_card_2025}OpenAI GPT-5 System Card.  (2025), https://openai.com/index/gpt-5-system-card/, Accessed: 2026-03-08

\end{thebibliography}
\end{document}